\documentclass[10pt,journal,compsoc]{IEEEtran}
\newcommand{\figref}[1]{Fig\onedot~\ref{#1}}

\newcommand{\ve}[1]{{\mathbf #1}} % for displaying a vector or matrix

\newcommand{\thickhline}{%
    \noalign {\ifnum 0=`}\fi \hrule height 1pt
    \futurelet \reserved@a \@xhline
}

\DeclareRobustCommand\onedot{\futurelet\@let@token\@onedot}
\def\onedot{\ifx\@let@token.\else.\null\fi\xspace}
\def\eg{\emph{e.g.}}

\def\ie{\emph{i.e.}}

\def\etc{\emph{etc}\onedot}

\usepackage{graphicx}
\usepackage{enumitem}
\usepackage{siunitx}
\usepackage{multirow}
\usepackage{xcolor}
\usepackage[normalem]{ulem}
\usepackage{color}

\usepackage[ruled]{algorithm2e} % For algorithms

\usepackage{booktabs} 

\usepackage{CJK}
\usepackage{amsthm,amsmath,amssymb}
\ifCLASSOPTIONcompsoc
  \usepackage[nocompress]{cite}
\else
  \usepackage{cite}
\fi
\ifCLASSINFOpdf
\else
\fi
\begin{document}
\title{PerMO: Perceiving More at Once from a Single Image for Autonomous Driving}

\author{Feixiang Lu, Zongdai Liu, Xibin Song, Dingfu Zhou, Wei Li, Hui Miao, Miao Liao, Liangjun Zhang, Bin Zhou,~\IEEEmembership{Member,~IEEE,}
        Ruigang Yang,~\IEEEmembership{Senior Member,~IEEE,}
        and Dinesh Manocha,~\IEEEmembership{Fellow,~IEEE}% <-this % stops a space
%\IEEEcompsocitemizethanks{\IEEEcompsocthanksitem M. Shell was with the Department
%of Electrical and Computer Engineering, Georgia Institute of Technology, Atlanta,
%GA, 30332.\protect\\
%% note need leading \protect in front of \\ to get a newline within \thanks as
%% \\ is fragile and will error, could use \hfil\break instead.
%E-mail: see http://www.michaelshell.org/contact.html
%\IEEEcompsocthanksitem J. Doe and J. Doe are with Anonymous University.}% <-this % stops an unwanted space
}


\markboth{Journal of \LaTeX\ Class Files,~Vol.~14, No.~8, July~2020}%
{Shell \MakeLowercase{\textit{et al.}}: Bare Demo of IEEEtran.cls for Computer Society Journals}



\IEEEtitleabstractindextext{%
\begin{abstract}
We present a novel approach to detect, segment, and reconstruct complete textured 3D models of vehicles from a single image for autonomous driving. Our approach combines the strengths of deep learning and the elegance of traditional techniques from part-based deformable model representation to produce high-quality 3D models in the presence of severe occlusions. We present a new part-based deformable vehicle model that is used for instance segmentation and automatically generate a dataset that contains dense correspondences between 2D images and 3D models. We also present a novel  end-to-end deep neural network to predict dense 2D/3D mapping and highlight its benefits. Based on the dense mapping, we are able to compute precise 6-DoF poses and 3D reconstruction results at almost interactive rates on a commodity GPU. We have integrated these algorithms with an autonomous driving system. In practice, our method outperforms the state-of-the-art methods for all major vehicle parsing tasks: 2D instance segmentation by 4.4 points (mAP), 6-DoF pose estimation by 9.11 points, and 3D detection by 1.37.
% In practice, for all vehicle parsing tasks, our method improves the state-of-the-art baseline by 4.4 (mAP) points for instance segmentation, 9.11 for 3D pose estimation and 1.37 for 3D vehicle detection.
%We use the reconstructed models for data-driven autonomous driving simulation systems, which use a combination of reconstructed models with synthetic datasets.
Moreover, we have released all of the source code, dataset, and the trained model on Github.

% (\textit{https://github.com/SA2020PerMo/PerMo}).
%We plan to release the code and the data set upon publication. 
\end{abstract}

}

\maketitle

\IEEEdisplaynontitleabstractindextext

% \fi
\IEEEpeerreviewmaketitle

\IEEEraisesectionheading{\section{Introduction}\label{sec:introduction}}

\begin{figure*}
  \centering
  \includegraphics[width=0.99\linewidth]{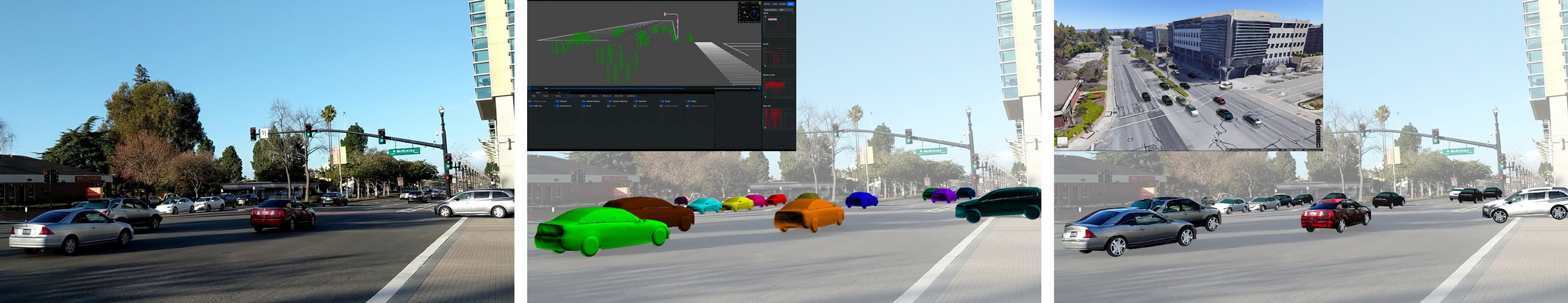}
  \caption{We highlight multiple perception results generated at an interactive rate from a single input image by our PerMO approach, including instance segmentation and 3D reconstruction of vehicle models. (left) single input image; (middle) 3D vehicle detection results used in an autonomous driving (AD) system; (right) reconstructed 3D textured models rendered from the same viewpoint. These 3D vehicle perception results computed using PerMO are used for various aspects of autonomous driving,  including  perception, planning, and simulation. }
  \label{fig:Teaser}
\end{figure*}
  
%\section{Introduction} 
%\modified{Autonomous driving (AD) is considered one of the most challenging problems in transportation and AI.  Overall, AD is an inter-disciplinary field and needs many advances in perception, planning, and control as well as successful integration with physical components. An important problem in the context of AD systems is detecting and analyzing moving objects, particularly vehicles, from perception data~\cite{arnold2019survey}. Active sensors such as LiDAR devices, which can provide robust and precise depth measurements of their surroundings, have been widely used in AD for perception. However, the high cost of LiDAR has resulted in use  of cheaper, commodity cameras. Stereo camera rigs are another way of obtaining the depth measurements based on projective geometry, but these methods usually suffers from the calibration issues between the two cameras. Therefore, many AD systems use single camera-based systems~\cite{arnold2019survey}. In general,  perception from a single camera is an important problem in both computer graphics (CG ) and computer vision (CV).} 

\IEEEPARstart{A}{utonomous} driving (AD) is considered one of the most challenging problems in transportation and Artificial intelligence (AI). Overall, AD is an inter-disciplinary field and needs many advances in perception, planning, and control as well as successful integration with physical components. An important problem in the context of AD systems is detecting and predicting moving objects, particularly vehicles, from sensor data~\cite{arnold2019survey}. Active sensors such as LiDAR devices, which can provide robust and precise depth measurements of their surroundings, have been widely used in AD perception. However, the high cost of LiDAR along with interference and jamming issues with laser pulses has resulted in the use of commodity cameras for AD. The latter include stereo camera rigs that can obtain the depth measurements based on projective geometry, but these methods usually suffer from calibration issues between the two cameras. Therefore, many AD systems use cameras or use cameras, Lidar, Radar, and other sensors in a hybrid combination \cite{arnold2019survey}. In general, camera perception is an important problem in both computer graphics (CG) and computer vision (CV).

In this paper, we mainly deal with the problem of 3D vehicle perception from a single image. In many ways, using a single image as an input to an AD perception system is simplest and simple case and can always be combined with other sensors.
Given a single image as an input, various methods have been developed for 2D bounding box detection (\eg,~\cite{ren2015faster,liu2016ssd}), instance-level segmentation (\eg,~\cite{maskrcnn:iccv2017}), 3D bounding box detection (\eg,~\cite{mousavian20173d}), and, more recently, 3D model fitting (\eg,~\cite{kundu20183d,yao20183d}). While these state-of-the-art methods have achieved good results for camera-based sensing, there are  many challenges with respect to handling complex scenarios with occlusion among the objects. Most prior approaches only parse the surrounding environment in the image space (\eg, 2D bounding box detection and instance segmentation), and do not provide 3D information about the detected objects. For 3D bounding-box detection, prior methods suffer from inaccurate 3D positions and poses. Furthermore, many autonomous driving simulation systems need reconstructed  textured vehicle models from a single image that can be useful for generating the training data for learning-based methods~\cite{li2019aads}.

% and cannot detect the vehicle action (\eg, opened door), which is essential for safety. 

% Given these demands, we designed and implemented a 3D vehicle perception system to detect, segment, and reconstruct complete textured 3D models from a single image. Our approach can improve performance, decrease overhead, and increase safety.
% Heavy occlusion is a major problem, which is difficult to detect/segment

 \subsection{Main Contributions}
 We present a novel approach for 3D vehicle perception from a single image. This includes new algorithms for object detection, segmentation, and reconstruction of metric 3D poses and 3D vehicle models. The key aspect of our method is combining traditional modeling techniques based on deformable representations with deep-learning-based approaches. Moreover, we present a new deep learning algorithm to solve the central problem of correspondence between the input image and the 3D model. We use a new part-based deformable representation of vehicles and an automatically generated training dataset that has dense per-pixel correspondences between the training images and the 3D models. The part-based model representation makes our approach robust to occlusions, which can be frequently observed in crowded traffic scenarios. To solve the correspondences problem, we first identify the part, then find pixel correspondences within each part. Given the correspondence, we use a traditional approach to reconstruct full 3D geometric and textured models by taking  advantage of the symmetry and geometric characteristics.

The novel aspects of our work include:

\begin{enumerate}
\item We present a novel approach for 3D vehicle perception from a single image that outperforms the state-of-the-art methods for all major vehicle parsing tasks: 2D instance segmentation by 4.4 points (mAP), 6-DoF pose estimation by 9.11 points, and 3D vehicle detection by 1.37. Our formulation also outperforms prior techniques for shape reconstruction.
\item In order to effectively handle the occlusion problem, we introduce a new part-level-based 2D/3D dense mapping. We develop a deformable model with specific parts definition and automatically generate a dense mapping dataset. The part-based vehicle dataset is also released to the public.
\item We present an end-to-end deep neural network to predict dense 2D/3D mapping and use that network to detect, segment, and reconstruct complete textured 3D models. 
%\item The proposed framework can be easily adopted in real applications for robotic and autonomous driving. We also hope that this work will bring the 3D perception in autonomous driving to a broad audience and inspire future research into this important problem.
\item Our approach has been integrated with autonomous driving perception systems and simulators. Our initial evaluations demonstrate that our proposed methods provide better perception results than other autonomous driving perception algorithms in instance segmentation, 6-DoF pose estimation, and 3D detection. The reconstructed 3D models and the dataset have also been used for data-driven autonomous driving simulation and vehicle behavior reasoning.
\end{enumerate}

%Experimental results demonstrate advantages of our approach with respect to object segmentation, part-level mapping, pose estimation, as well as deformable model reconstruction.

%\modified{Our algorithm offers almost interactive performance and is a linear function of the number of cars in the image. For a $3384 \times 2710$ image containing more than 10 vehicles, the average processing time is about $0.2$s seconds for reconstruction and $0.12$s for pose and shape estimation on an NVIDIA Tesla P40 GPU.
%We highlight multiple applications of our approach to AD in Fig.~\ref{fig:Applications_AD} . First, the 6-DoF vehicle pose (or 3D bounding box) result is used for reliable vehicle tracking and prediction. Second, when applied to a video sequence, our approach captures traffic dynamics that are employed in a simulator to test self-driving navigation policies against real-world scenarios. Third, our reconstructed vehicle 3D model is rendered in a new virtual set or mixed with real images to generate a large amount of ground truth data for training neural networks. Fourth, based on the parts segmentation, the part state (e.g., the left-door is open) of the vehicle is determined based on a simple binary classifier. According to the state information, further action of the vehicle can be effectively predicted.}

\begin{figure}
  \centering
  \includegraphics[width=1.0\linewidth]{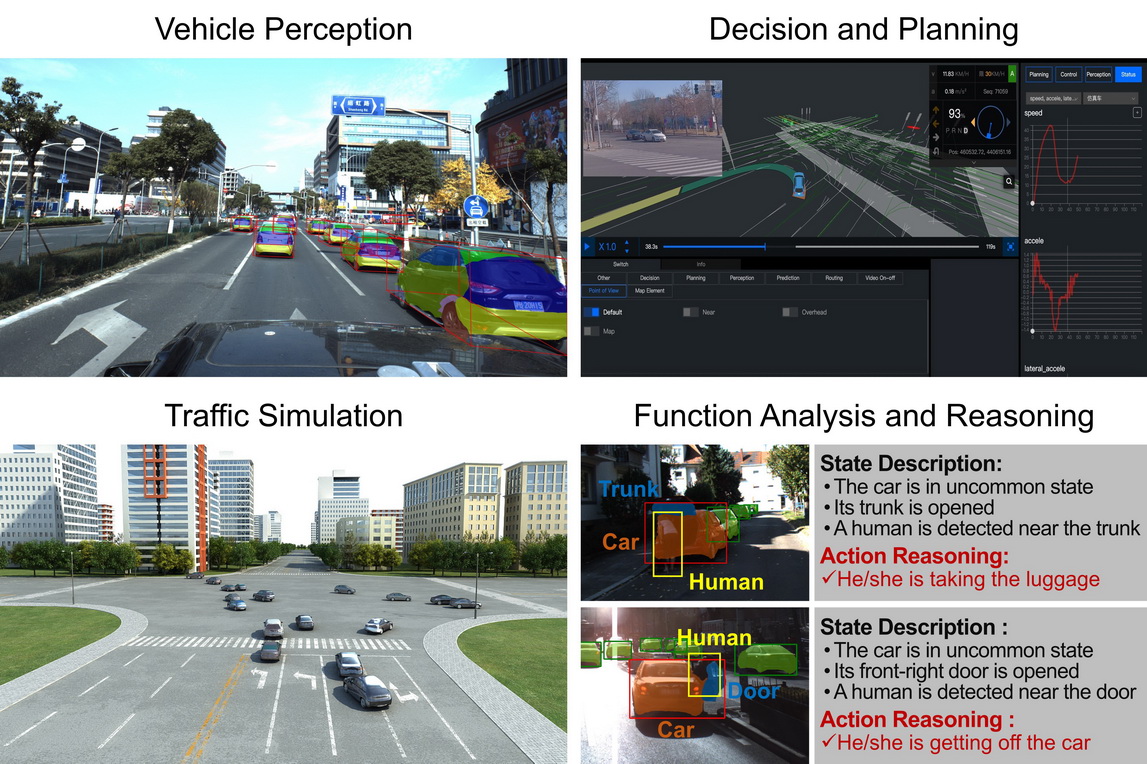}
  \caption{PerMO has been integrated into current AD systems and used to improve the performance of several components including perception, decision and planning, data-driven simulation, and function analysis and scene reasoning. 
  }
  \label{fig:Applications_AD}
\end{figure}

Our algorithm offers almost interactive performance on a PC and the running time is a linear function of the number of cars in the image. For a $3,384 \times 2,710$ image containing more than $10$ vehicles, the average processing time is about $0.2$s seconds for reconstruction and $0.12$s for pose and shape estimation on an NVIDIA Tesla P40 GPU. We highlight multiple applications of our approach to AD in Fig.~\ref{fig:Applications_AD}. First, the 6-DoF vehicle pose (or 3D bounding box) result is used for reliable vehicle tracking and prediction. Second, the detected vehicles' 3D pose and shape information are sent to AD decision and the planning module to generate feasible vehicle control commands. Third, we apply our approach to a video sequence and extract the traffic dynamics from multiple frames. This information is used in an autonomous driving simulator to evaluate self-driving navigation policies.
% Third, our reconstructed vehicle 3D model is rendered in a new virtual set or mixed with real images to generate a large amount of ground truth data for training neural networks. 
Fourth, we use parts segmentation to determine the part state (e.g., the left-door is open) of the vehicle and use that state information to predict further action of the vehicle.
%According to the state information, further action of the vehicle can be effectively predicted.

The remainder of this paper is structured as follows. Sec.~\ref{sec:Related_work} reviews related work. Sec.~\ref{sec:overview} introduces some of the perception challenges with respect to autonomous driving and gives an overview of our approach. We introduce our self-designed part definition and deformable vehicle representation in Sec.~\ref{sec:3D_car_representation}. Based on the vehicle template, we automatically generate a dataset with 2D/3D dense mapping in Sec.~\ref{sec:dataset}. The learning-based vehicle parsing and geometric reconstruction are presented in Sec.~ \ref{sec:2d_parsing} and Sec.~\ref{sec:shape_pose}. We compare our approach with state-of-the-art methods on KITTI and ApolloCar3D datasets and discuss the applications to AD in Sec. \ref{sec:Experimental_Results}.
%%%%%%%%%%%%%%%%%%%%%%%%%%%%%%%%%%%%%%%%%

\begin{figure*} [!htbp]
  \centering
  \includegraphics[width=0.98\linewidth]{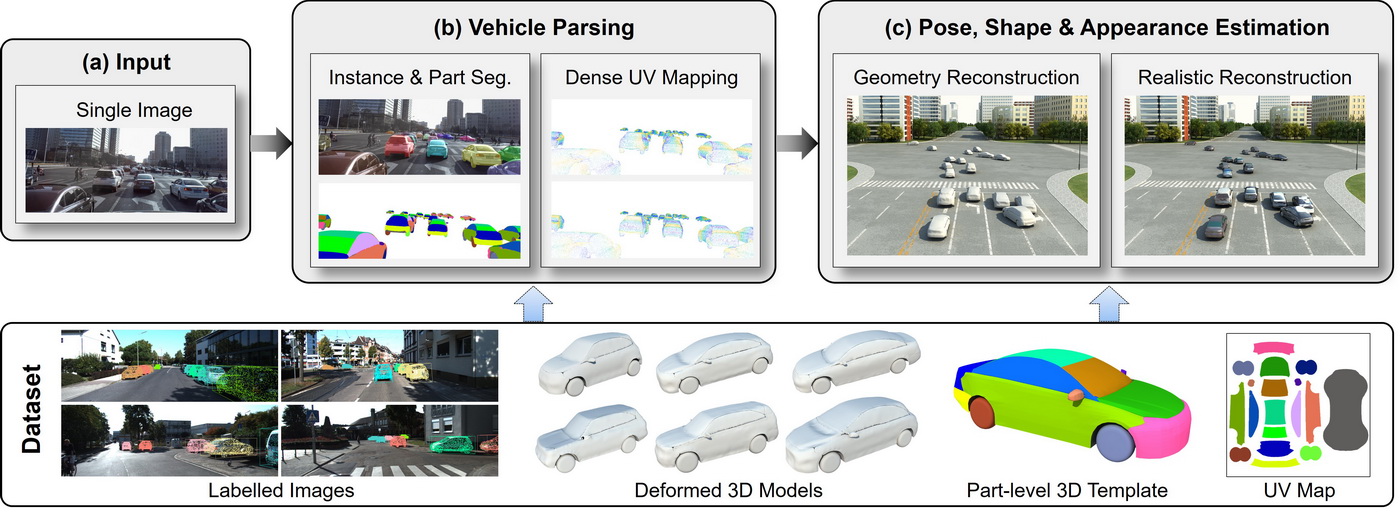}
  \caption{Overall pipeline of our approach: given a single image (a), a two-step framework is used to fully parse and reconstruct all vehicles at once. The first step (b) is the complete vehicle parsing step, which uses an end-to-end deep network to predict instance and part-level segmentation as well as inner-part dense mapping between the input image and the 3D deformable model. With such dense information, a more accurate reconstruction based on classical optimization techniques is used in the following step (c). We use a  gradient-based optimization method to refine the 6-DoF pose and the shape and reconstruct the realistic appearance. Finally, we automatically generate the dataset of dense per-pixel correspondences between the training images and the deformable 3D models. The dataset is used for the network training and the 3D vehicle pose estimation and reconstruction.
}
  
  \label{fig:System_overview}
\end{figure*}

\section{Related works}
\label{sec:Related_work} 
%%%%%%%%%%%%%%%%%%%%%%%%%%%%%%%%%%%%%%%%%

%3D reconstruction has been a long-standing problem in computer graphics and computer vision. In terms of 3D reconstruction of static scenes, there are a large number of previous works in the literature. For static background reconstruction, we refer readers to follow the following research \cite{wu2011visualsfm, scaramuzza2011visual,fraundorfer2012visual,schonberger2016structure}. Dynamic object reconstruction as a basic topic in computer vision and graphic field has been studied for a quite long time as well. Here, we only review works related to rigid object reconstruction while approaches related to non-rigid object reconstruction are out of our scope in this paper. 

Perception is one of the most important modules for AD and has been well studied in the literature~\cite{arnold2019survey, rahman2019recent}. Different approaches have been developed \cite{zhou2018voxelnet, zhou2019iou, nabati2019rrpn, li2019stereo} based on different sensors (e.g., LiDAR, Radar, camera, etc.).  In this section, we limit our discussion to methods that also use a single image for AD scenarios. At a broad level, we can divide the prior techniques into three categories: 2D Bounding Box (BBox) detection and semantic parsing, 3D pose estimation or BBox detection, and full 3D model reconstruction. 
%We review related works within these three areas in the following sub-sections. 

\subsection{2D BBox Detection and Semantic Parsing}
Semantic parsing is a useful technique for understanding the structures of objects in an image and is also related to semantic segmentation~\cite{chen2017deeplab}.
%before further modeling those objects.
%However, semantic segmentation is a major research area in its own right~\cite{long2015fully, chen2017deeplab} and it is not sufficient to describe the relationship of different objects in the scene. 
Instance level semantic segmentation is an active area of research that deals with parsing objects separately, which is needed for 3D object perception from a single image. A well-known algorithm for instance semantic segmentation~\cite{long2015fully} uses a fully convolutional network (FCN) to parse a single image into different semantic subjects. Recent advances in 2D object detection using Fast/Faster-RCNN~\cite{girshick2015fast,ren2015faster} have resulted in better deep learning solutions (such as instance segmentation Mask-RCNN \cite{maskrcnn:iccv2017}) that work well on well-known datasets~\cite{cordts2016cityscapes}.

%Targeting human body parsing,~\cite{li2018multi} present results on multiple human body segmentation at once. By parsing human bodies, positions of each body parts and their relation can be easily understood, and this in turn can be used to compute poses/actions. There is also extensive work on extending these ideas to 3D point clouds~\cite{qi2017pointnet, qi2017pointnet++}. 

\subsection{3D BBox Detection and Pose Estimation}
It turns out that the planning and control modules of an AD system need more information than that provided by 2D BBox or instance segmentation. Therefore, there has been some work on recovering 3D information such as 3D bounding boxes or poses and shapes. In terms of traffic entities and vehicles, some of the well-known methods for predicting a 3D bounding box are based on key-points detection~\cite{song2019apollocar3d}. Key-points logically represents the vehicle parts, e.g., centers of the wheels or corners of doors. Given the detected key-points in an input image, the 3D bounding box and the 3D shape of a vehicle can be extracted using a PnP solver~\cite{lepetit2009epnp}. However, the performance of these methods degrades when parts of the vehicles are occluded, which leads the pose estimation problem to be constrained by limited visible key-points. 

%By translating the key-points into vehicle parts and then encoding them into an end-to-end network, the method of~\cite{geng2018part} could output part-level segmentation and directly estimate 3D bounding boxes or poses of vehicles with better robustness and performance.
There is some work on utilizing the dense part information to improve the robustness of pose estimation. Representative works including DenseReg~\cite{alp2017densereg} and Densepose~\cite{DensePose:CVPR2018} have advanced the state-of-the-art for face alignment and human pose estimation. This success comes from the fact that the dense correspondences between 2D image pixels and object surfaces are first predicted and then used to estimate the alignment or pose. Our approach is also motivated by such techniques.

\subsection{Full 3D Model Reconstruction}
For some applications, \eg{} obstacle avoidance, an object-level 3D bounding box may be sufficient. However, fully reconstructing the 3D object is important for AD, including collision avoidance and path planning in dense scenarios. With the development of deep CNNs, researchers have been able to achieve impressive results for semantic segmentation and object detection with supervised or weakly supervised methods. These works represent an object as a parameterized 3D bounding box~\cite{fidler20123d, poirson2016fast, xu2018multi}, a coarse wire-frame skeleton~\cite{li2017deep,zia2015towards}, a voxel-based representation~\cite{choy20163d, xiang2015data}, or select from a small set of exemplar models~\cite{chabot2017deep, mottaghi2015coarse, guo2008matching, ortiz2016automatic}. In \cite{wang2019normalized}, Wang et al. proposed to estimate the 6D pose and dimensions of unseen object instances in an RGB-D image. In addition, they introduce Normalized Object Coordinate Space (NOCS)—a shared canonical representation for estimating the pose for unseen object instances. While these works have substantially advanced the state-of-the-art, they can still result in inaccurate shapes and empty or rough appearances when creating  high-fidelity reconstructions from real traffic scenarios. Instead of using deep networks to fully reconstruct the 3D vehicle model, we use a network to improve the robustness of the dense mapping estimation and then reconstruct the accurate shape and texture using a separate pipeline. 

%In contrast to direct meshing on point cloud, 
Techniques based on deformable model regression  have been used to reconstruct objects corresponding to complicated shapes. PCA-based deformable modeling has been successfully employed to model human faces~\cite{Blanz:1999:MMS:311535.311556}, and extended to the human body~\cite{guan2009estimating,zhou2010parametric}.
%Inspired by this idea, similar methods have been proposed to model vehicles.
3D-RCNN~\cite{kundu20183d}, DeepMANTA~\cite{chabot2017deep} and DirectShape \cite{wang2019directshape} are state-of-the-art techniques that combine 3D models and 2D object detection for 3D object shape reconstruction. Based on these prior works, we extend the deformable modeling \cite{lopez2011deformable} approach by integrating the semantic parsing results as geometric constraints to generate high-fidelity vehicle models.

\section{Overview}
\label{sec:overview}
%%%%%%%%%%%%%%%%%%%%%%%%%%%%%%%%%%%%%%%%%%%%%%

\begin{figure}[t]
 \includegraphics[width=1.0\linewidth]{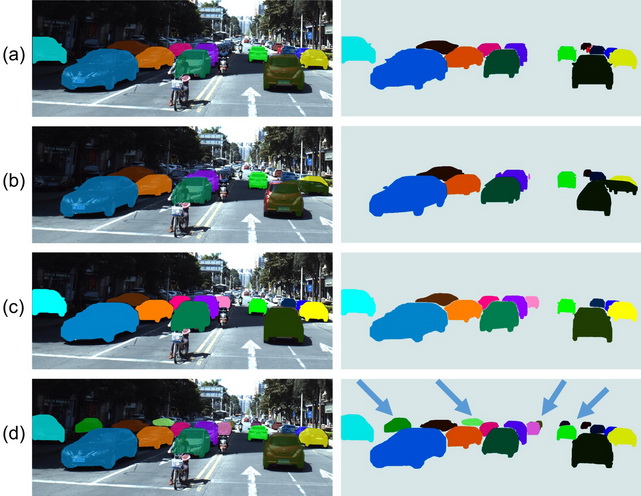}
%\caption{Segmentation examples in ApolloCar3D dataset for highlighting the advantage of our method to handle the occlusion cases.}
\caption{Instance segmentation on a heavy occlusion scenario. The task of instance level of semantic segmentation is to detect each distinct object of interest appearing in an image, e.g. each vehicle should be detected, and the pixels of the vehicle should be further labeled with distinct vehicle instance ID. (a) Mask-RCNN~\cite{maskrcnn:iccv2017}; (b) ApolloCar3D~\cite{song2019apollocar3d}; (c) PANet~\cite{liu2018path}; (d) our method. Note that Mask-RCNN, ApolloCar3D, and PANet segment out $14$, $11$, and $12$ vehicle instances, respectively. In contrast, our method can segment out $18$ vehicles. The main improvement comes from the occluded vehicles, which demonstrates the effectiveness of our part-based approach.}
\label{fig:seg-zoomin}
\end{figure}

%In this section, we give an overview of our approach for 3D vehicle perception from a single image.
3D perception is essential for AD in terms of functionality, performance, and safety. In this paper, we focus on the problem of 3D vehicle parsing from a single image, which is challenging, especially for handling complex scenarios with heavy occlusion among the objects. A typical city driving scenario is shown in Fig.~\ref{fig:seg-zoomin}, where there are vehicles stopped at the stop light. From the view-point of the ego vehicle, many of the other vehicles are partially occluded. Existing methods that rely on feature extraction from the entire vehicle cannot effectively segment vehicles or detect the poses due to the occluded regions, yielding incorrect instance mask segmentation and inaccurate 3D vehicle poses. We explicitly exploit parts information of the vehicle model (\eg{}, headlights, doors, and trunk), which provides strong priors and constraints to handle the occlusion problem. Furthermore, in contrast to other approaches that recover 2D vehicle information mainly using 2D image pixel semantic label, we combine the 3D shape and geometry information of vehicle models together into the learning process.

% , it has been explored in other applications corresponding human body segmentation and face recognition. Unfortunately, there is not AD-oriented part-level vehicle database that can be used to address this problem.}
% In this section, we present an overview for our approach for 3D vehicle parsing from a single image produce in the presence of severe occlusions. Our approach combines the strength of the deep learning and the model-based techniques. 
% We learn a model of per-pixel dense mapping between 2D image and the 3D deformable vehicle model surface. This mapping is defined on top of the middle-level functional parts of vehicles, such as the wheels, doors, \etc{}. We establish the relationship between each part by representing them with uniform UV-map coordinates. 

As illustrated in Fig.~\ref{fig:System_overview}, our approach consists of two steps for fully parsing and reconstructing all vehicles at once. The first step is the vehicle parsing, which uses an end-to-end deep network to predict instance and part-level segmentation as well as inner-part per pixel dense mapping between 2D image and the 3D deformable vehicle model surface. The key for this step is to learn a model of the dense mapping, which is defined on top of the middle-level functional parts of vehicles, such as the wheels, doors, \etc{}. To do so, we unfold a 3D model surface and construct the UV mapping for all its parts. We train a network of dense mapping to predicate the part and UV coordinates for 2D image pixels, so we can map them back to 3D model surface. The network extracts features from the input image using Region Proposal Network (RPN) and ROI-Align pooling and feeds the resulting RoI features into multi-task heads to obtain the instance-level segmentation, part-level segmentation, vehicle type and the dense U/V coordinates. At the end of the first step, we add a simple and effective network module, which uses the learned 2D/3D dense mapping and determines the initial position for each vehicle. The second step of our approach is to determine a more accurate vehicle pose and shape, and reconstruct the completed textured module. In our approach, we propose a gradient-based optimization method to refine the 6-DoF pose and the shape, and reconstruct the realistic appearance. 

In our approach, we further propose a novel automatic dense map dataset  generation for the training process. In order to generate dense mapping data between the pixels of 2D images and the UV coordinates of the 3D vehicle representation, we use a deformable 3D model as the bridge to deform to various target 3D vehicles models and account for the model variation in real world. By deformation, the U/V mapping information of the source deformable model can be automatically transferred to the target models.

% \modified{CAN YOU DEFINE THE PROBLEM AND DENSE MAPPING MORE FORMALLY? THE CURRENT OVERVIEW DON"T REALLY GIVE AN OVERVIEW OF YOUR APPROACH.} 
% We learn a model of per-pixel dense mapping between 2D image and the 3D deformable vehicle model surface. This mapping is defined on top of the middle-level functional parts of vehicles, such as the wheels, doors, \etc{}. We establish the relationship between each part by representing them with uniform UV-map coordinates. We handle the occlusion problem for these tasks by effectively considering part-level correspondences. 
% {CAN YOU ELABORATE MORE WHY OCCLUSION OCCURS FREQUENTLY, ESP. IN DENSE SETTINGS WITH A FIGURE. YOU SHOULD ALSO DEFINE OR EXPLAIN WHAT DOES PART-LEVEL CORRESPONDENCE MEANS (e.g. WITH A FIGURE) AND HOW IT OVERCOMES OCCLUSION (done)}

Our approach is related to recent methods that implicitly or explicitly estimate 3D pose and geometric models for vehicles~\cite{kundu20183d,yao20183d,song2019apollocar3d}. Compared with these algorithms, our formulation produces much better segmentation and pose accuracy with our parts-level 3D vehicle models. Our formulation also generates an explicit CG model that can be combined with other CG models and manipulated as a geometric shape, which is rather difficult with implicit representation. 

\section{Part-level Deformable Vehicle Representation} 
\label{sec:3D_car_representation}
%%%%%%%%%%%%%%%%%%%%%%%%%%%%%%%%%%%%%%%%%

In this section, we present our part-based deformable vehicle representation that is used for accurate 3D reconstruction from a single image. The notion of decomposing an object into multiple parts with a global layout representation is widely used for object detection~\cite{felzenszwalb2010object} and 3D pose estimation~\cite{zia2013detailed}. Compared to a single template model, the main advantage for part-based model representation is the ability to account for variations in object shape and robustness to \textit{partial occlusion}.

\begin{figure}
  \centering
  \includegraphics[width=1.0\linewidth]{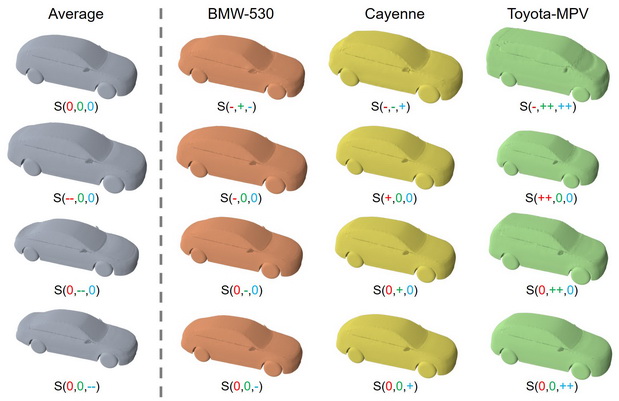}
  \caption{Our deformable vehicle representation and its application to different types of vehicles, e.g. sedan (BMW-530), SUV (Cayenne), and MPV (Toyota-MPV). The left corner of this figure highlights the mean model of our representation. The other images correspond to the representation with different coefficients including the top 3 principal components. For simplicity, ``0'' represents the component coefficient of the mean model, and ``+'' and ``-'' represent the increase and decrease of the corresponding coefficients, respectively. The result shows that our deformable model is general for deforming to different shapes of models. }
  \label{fig:PCA_Basis}
\end{figure}

\begin{figure*}
  \centering
  \includegraphics[width=1.0\linewidth]{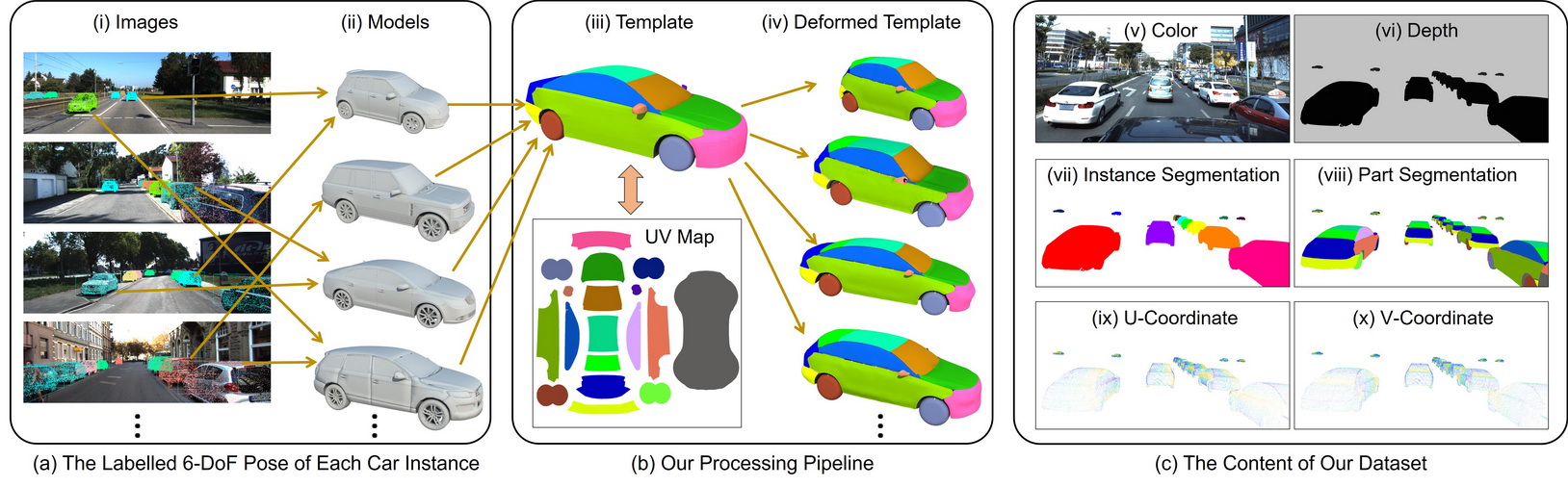}
  \caption{Overall pipeline for automatically generating our training dataset:  (a) All of the instances in ApolloCar3D and KITTI with labeled 3D models and 6-DoF poses. In (b), the image in the top left corner illustrates our deformable vehicle representation, and the unfolded UV-map is displayed below. Several deformed vehicle models together with ground truth labels are shown on the right side of (b). The generated ground truth labels are shown in (c), including instance labels, parts labels, 6-DoF poses, depth, and U/V-Coordinate maps.}
  \label{fig:AnnotatedDataset}
\end{figure*}

Part-based deformable models have been used in CV and CG. However, no good methods are known for general and diverse types of vehicles. Moreover, it is non-trivial to define and create a proper part-based vehicle representation for \textit{2D tasks} (detection and segmentation) as well as \textit{3D tasks} (pose and shape reconstruction).
There is not much prior work on ready-to-use part definition of vehicles for instance segmentation or 3D reconstruction. One approach is to directly inherit the definition of components  from  industrial designed CAD models. In \cite{geng2018part}, each car was decomposed into $70$ exterior parts, including \textit{large parts} such as the front door and the roof, and \textit{small parts} such as the door handle and the car logo. However, these representations are not suitable for segmentation or 3D reconstruction applications. For example, some thin or small parts are not clearly visible in an image.

%%%%%%%%%%%%%%%%%%%%%%%%%%%%%%%%%%%%%%%%%%%%%%%%%%%%%%%%%%
% \subsection{Vehicle Geometry with Parts Re-definition}
\subsection{3D Vehicle Part-Level Representation}
%%%%%%%%%%%%%%%%%%%%%%%%%%%%%%%%%%%%%%%%%%%%%%%%%%%%%%%%%%
We have designed our 3D vehicle part representation based on the following goals. \textit{1) integrality}: a part should correspond to one or more complete components of the 3D vehicle model; \textit{2) universality}: a part should be covered in most of the sub-classes belonging to the same category; \textit{3) uniqueness}: each part should have its own characteristics distinct from other parts. Specifically, we represent the vehicle model with $N$ (\ie{} 18) small components as a trade-off between model representation precision and generality. The detailed information for each part is given in the supplementary material. As shown in \figref{fig:AnnotatedDataset} (b), some components are derived from the original CAD model of a vehicle directly, \ie{} the front window, chassis, roof, etc. For some small parts, we group them into one component. For instance, we group the headlights, the grille, the front bumper, and the  vehicle logo as ``the front face'' because these components vary with different brands of vehicles. In \figref{fig:AnnotatedDataset}, ``the front face'' is illustrated in pink. In addition, in order to generally cover different types of vehicles (\eg{}, hatchback and sedan), we merge the front and rear doors together as a ``door'' component. 

Based on the part representation, we use a 3D vehicle model representation for model deformation. In order to capture the geometry details, we set the mesh resolution of the vehicle model as $0.018$ \si{\meter} and the final template model consists of $172,642$ vertices and $341,184$ faces. As shown in Fig.~\ref{fig:AnnotatedDataset} (b), we  unfold the 3D model to generate a UV-map with a commercial tool, Unfold3D \footnote{http://www.polygonal-design.fr/e\_unfold}, where each color denotes an individual part. The constructed vehicle model is included in the supplementary material. 

% Furthermore, we compute the corresponding UV-coordinate for each vertex to build the 2D/3D dense mapping.
% \begin{equation}
%     E_s(\mathcal{T}, \mathcal{M}) = \sum_{x,y} \|\mathcal{I}_{\mathcal{S}}(x,y)  - \mathcal{I}_{\mathcal{M}}(x,y) \|^2.
% 	\label{equ:termSil}
% \end{equation}

% Given these requirement, we designed our 3D vehicle \modified{model representation together with part labels.}

%%%%%%%%%%%%%%%%%%%%%%%%%%%%%%%%%%%%%%%%%%%%%%%%%%%%%%%%%
\subsection{Deformable Vehicle Representation}
\label{subsec:deformable_template}
%%%%%%%%%%%%%%%%%%%%%%%%%%%%%%%%%%%%%%%%%%%%%%%%%%%%%%%%%

To efficiently transfer the part label information from the template model to other target models, we build a PCA basis. First, we deform the template model to $106$ CAD vehicle models (ApolloCar3D \cite{song2019apollocar3d} provides $78$ 3D vehicles and the remaining $28$ models are manually constructed by artists). Then, all the deformed models share the same order of vertices and faces as the template model. Finally, a PCA basis of $517,926 \times r$ is constructed based on the deformed $106$ models, where $r$ is the number of top principal components.

With this PCA basis, any new 3D vehicle model $\mathcal{M}(s)$ can be represented as a linear combination of $r$ principal components with coefficient $s=[s_1, s_2, ..., s_r]$ as:
\begin{equation}
    \mathcal{M}(s) = \mathcal{M}_0 + \sum_{k=1}^{r}s_k\delta_kp_k,
	\label{equ:pcamodel}
\end{equation}
where $\mathcal{M}_0$ corresponds to an average model, $p_k$ and $\delta_k$ are the principal component direction and corresponding standard deviation and $s_k$ is the coefficient of the $k_{th}$ principal component.

In order to generate the PCA model, we first align the CAD vehicle models with the average template model. We use a bidirectional deformation scheme to deform the 3D template to the target CAD models (more details are given in the appendix). From \figref{fig:PCA_Basis}, we see that our PCA basis is quite general and can generate both the SUV and the van, which are quite different from our base model representation. In this manner, we use a general part-based deformable representation for the vehicles.

\section{2D/3D Dense Mapping}
%%%%%%%%%%%%%%%%%%%%%%%%%%%%%%%%%%%%%%%%%%%%%%%%%%%
\label{sec:dataset} 

In this section, we introduce our novel algorithm to automatically generate 2D/3D dense mapping datasets based on the deformable 3D representation (Sec.~\ref{sec:3D_car_representation}) and labeled vehicles with 6-DoF poses (Sec.~\ref{subsec:data_labeling}). For 2D car instances in every camera image of our dataset, we use multiple automatically generated ground-truth labels including: instance-level segmentation, part-level segmentation, depth, and U/V coordinates (Fig.~\ref{fig:AnnotatedDataset} (c)).

%In this section, we present our deep learning based method for predicting the dense mapping from 2D image to 3D surface model, the part level segmentation and vehicle pose simultaneously. Before presenting our network in detail, we will introduce our novel approach for dense mapping and other training data generation. 

%%%%%%%%%%%%%%%%%%%%%%%%%%%%%%%%%%%%%%%%%%%%%%%%%%%
\subsection{Labeled Dataset Generation}
\label{subsec:data_labeling}
%%%%%%%%%%%%%%%%%%%%%%%%%%%%%%%%%%%%%%%%%%%%%%%%%%%
Our dense mapping data comes from two published and widely used AD-oriented datasets: ApolloCar3D and KITTI. ApolloCar3D provides both the 3D CAD models and 6-DoF pose labeling for each car instance. Therefore, we can automatically generate the dense mapping 2D/3D data through the proposed pipeline (Sec.~\ref{subsec:data_pipeline}). The KITTI dataset only provides 3D bounding box annotation, so we manually label the KITTI dataset with 3D vehicle models and corresponding 6-DoF poses.

In order to efficiently label the KITTI dataset, we design a 3D vehicle annotation tool (see Appendix).
% Fig.~\ref{fig:labellingTool}
The pose information for a vehicle contains the rotation and translation parameters between the 3D vehicle model coordinate frame and camera coordinate frame. To label the poses of vehicles in the dataset, \textit{28} industrial grade vehicle CAD models are used, including five vehicle classes: \textit{coupe}, \textit{hatchback}, \textit{notchback}, \textit{SUV}, \textit{MPV}. To label the poses of vehicles in the KITTI dataset accurately, we use point cloud and 3D bounding box information  in the labeling process and enforce two constraints. First, for a vehicle in one image of the KITTI dataset, we first choose a suitable 3D vehicle model according to its shape type, then compute the corresponding 3D bounding box from the point cloud by modifying the rotation ($\alpha, \beta, \gamma$) and translation ($t_x, t_y, t_z$) parameters. The transformed 3D model should be aligned with the annotated 3D bounding box in KITTI, which is the first constraint.   
% WHAT IS THE EXACT CONSTRAINT, THIS IS NOT CLEAR (done). 
Further, rotation and translation need to be fine-tuned to guarantee that the re-projection error of the 3D vehicle model in the 2D image is below a small number (\eg{}, 3) of pixels. In summary, we manually label the training set of the KITTI dataset, which contains $6,871$ images and $33,747$ car instances in total. The labeling tool and labeled data (3D vehicle model and corresponding 6-DoF pose) are also released on the Github.

%To further improve the diversity of the training data, we provide the labeling results of the poses of vehicles for the KITTI dataset~\cite{geiger2012we}, and Fig.~\ref{fig:labellingTool} shows the 3D vehicle labelling tool we designed. The pose information for a vehicle contains the rotation and translation parameters between the 3D vehicle model coordinates and camera coordinates; to label the poses of vehicles in dataset, $36$  industrial grade vehicle CAD models were made. To label the poses of vehicles in the Kitti dataset accurately, the provided point cloud and 3D bounding box information are utilized in the labeling process and two constraints are enforced. First, for a vehicle in one image of the Kitti dataset, we first choose a suitable 3D vehicle model and put it in the corresponding 3D bounding box in corresponding point cloud by modifying the rotation ($R$) and translation ($T$) parameters. This is regarded as the first constraint. Further, $R$ and $T$ need to be fine-tuned to guarantee that the re-projection error of the 3D vehicle model in 2D image is below 5 pixels.

\begin{figure*}
  \centering
  \includegraphics[width=0.98\linewidth]{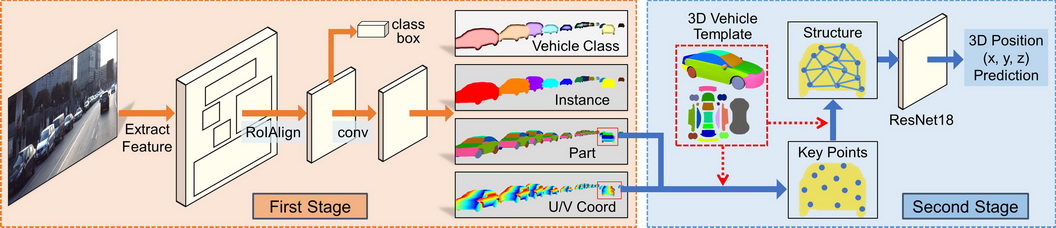}
  \caption{Our two-stage vehicle parsing network: The first stage takes a single image as input, and computes the vehicle type classification, instance-level and part-level segmentation, and dense UV coordinates. This approach of computing multiple outputs simultaneously helps with the overall computation. By using the novel dense UV mapping, the second stage predicts the 3D positions of the vehicles. }
  \label{fig:PosePrediction}
\end{figure*}

%%%%%%%%%%%%%%%%%%%%%%%%%%%%%%%%%%%%%%%%%%%%%%%%%%%
\subsection{Automatic Data Generation with 2D/3D Mapping}
\label{subsec:data_pipeline} 
%%%%%%%%%%%%%%%%%%%%%%%%%%%%%%%%%%%%%%%%%%%%%%%%%%%
To perform the learning-based fine-grained vehicle parsing, four types of ground truth labels are automatically generated: instance-level segmentation labels, part-level class labels, a depth map, and UV coordinates that correspond to 3D vertices of the vehicle's deformable representation. The instance-level label is a binary mask that represents whether a pixel belongs to the object in the foreground or the background, and this information is provided from the ApolloCar3D and KITTI datasets. Part-level class labels are consecutive integers for representing different parts of a 3D model. The depth maps are floating values, showing the depth for the pixel in the 3D space. The UV coordinates are floating values between 0 and 1, which represent the correspondence location of a pixel in the UV-space unfolded from the 3D model surface.

In order to generate dense mapping data between the pixels of 2D images and the UV coordinates of the 3D vehicle representation, we use the 3D models as the bridge because both the ApolloCar3D dataset and the KITTI dataset have labeled 3D target vehicle models and ground truth 6-DoF poses. Meanwhile, we use our PCA computation (Sec.~\ref{subsec:deformable_template}) to generate the  deformable  vehicle representation, its 2D UV-space for each part, and the dense 3D-to-3D mapping between the deformable template model and the labeled 3D CAD model.

We summarize the whole process in the following steps, which are highlighted in Fig.~\ref{fig:AnnotatedDataset}. First, we unfold or map the 3D surface of the template model to the 2D UV-space. Then, a non-rigid deformation is applied to deform the template model to the corresponding target model. Meanwhile, the part class information and the UV-coordinates are transferred to the target model simultaneously. Finally, all the ground truth labels including UV coordinates for the image pixels are generated by rendering the 3D model into the 2D image. The occlusion between the cars is handled automatically by enabling depth culling during rendering, as each image pixel has the depth information.

In summary, we build our training dataset, which contains more than $12,000$ labeled images with ground-truth 2D/3D dense mappings of $100,000$ vehicles (ApolloCar3D for $5,200$ images with $60,000$ instances; the training set of the KITTI dataset for $6,871$ images with $33,747$ instances).

\section{Learning based Fine-grained Vehicle Parsing}
\label{sec:2d_parsing}
%%%%%%%%%%%%%%%%%%%%%%%%%%%%%%%%%%%%%%%%%%%%%%%%%%%

Given the 2D/3D dense mapping dataset, we propose an effective deep CNN network to perform fine-grained vehicle parsing for an input single image. To obtain more information about vehicles from a single image, vehicle poses and shapes are predicted. In this paper, we use instance-level and part-level segmentation as well as vehicle types to obtain such information. Since UV information and part-level segmentation contain more effective information, we propose using this information to construct a keypoints-based graph to obtain vehicle positions. Hence, our network consists of two stages: the first stage is designed for instance-level and part-level segmentation, vehicle type classification, and UV prediction, and the second stage is designed for predicting the vehicle positions.

In summary, the outputs of the network include instance-level segmentation, part-level segmentation, vehicle type classification, dense UV coordinates, and preliminary 3D vehicle positions. In this section, we present our network architecture and the training settings in detail.

\subsection{Network Architecture}
Fig.~\ref{fig:PosePrediction} illustrates the architecture of our network, which is divided into two stages. In the first stage, using a color image as input, the network first extracts effective features, which are the inputs of the Region Proposal Network (RPN). Note that commonly used backbones such as Resnet18, Resnet50, VGG, etc. can be used in feature extraction, and we use Resnet50 empirically. Next, ROI-Align pooling is used to obtain ROI features, which are fed into multi-task heads. Then, we propose 
using four sub-branches to obtain the instance-level segmentation, part-level segmentation, vehicle type, and the dense U/V coordinates. More details are presented in Sections~\ref{sec:instance_seg}, ~\ref{sec:part_uv}, and~\ref{sec:type_prediction}. In the second stage, we design a simple and effective network to obtain the 3D vehicle position of each vehicle, which contains a backbone for feature extraction and fully connected convolution for position regression. Specially, for each detected vehicle instance, a-key-point based graph is first constructed using the predicted segmentation and U/V information as well as  the designed 3D Vehicle Template. Then, the graph is fed into a backbone (Resnet18) for feature extraction. Finally, a fully connected convolution is used to regress the 3D vehicle position using the obtained feature as input. More details are provided in Section~\ref{sec:keypoint_pose}.

\subsection{Instance Segmentation}\label{sec:instance_seg}
In this section, we provide more details about the sub-branch of instance segmentation. Similar to Mask-RCNN~\cite{he2017mask}, the sub-branch outputs a $\mathit{Km^2}$ dimensional binary mask for each ROI aligned feature, where $\mathit{K}$ is the number of classes and $\mathit{m}$ is the resolution. Specifically, inspired by~\cite{he2017mask}, we feed a 14$\times$14 ROI aligned feature map to four sequential 256-d 3$\times$3 convolution layers. A 2$\times$2 deconvolution layer is used to up-sample the output to 28$\times$28. Finally, we define the $\mathit{L_{mask}}$ as the average of per-pixel sigmoid cross entropy loss.

\begin{figure*}
  \centering
  \includegraphics[width=0.98\linewidth]{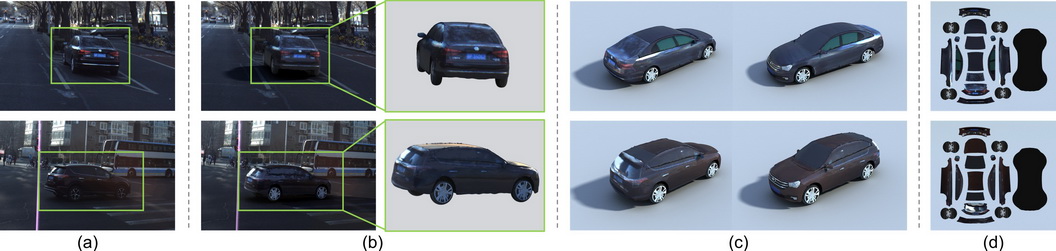}
  \caption{Appearance extraction from input images: (a) the input images; (b) images with reconstructed 3D vehicle models. The models are also rendered into a plain background in the green boxes; (c) the models are rendered into different views; (d) the reconstructed texture maps.}
  \label{fig:CompareKF&DF}
\end{figure*}

\subsection{Part Segmentation and Dense UV Prediction}\label{sec:part_uv}
In the sub-branch, we use a fully convolutional network (FCN) and divide the prediction of dense correspondences between pixels $p_i \in \mathcal{I}$ in the image space and 3D points $v_j\in \mathcal{M}$ on the surface mesh into a classification step and a regression step; a coarse-to-Fine strategy is used in these steps. First, the network makes a coarse estimate of  the surface coordinates, which classifies a pixel as belonging either to the background or to one of several region parts. Then, a regression network is used to predict the exact coordinates of pixels in the region part. We use a local two-dimensional coordinate system  to parametrize each part, which identifies the position of any node on the surface.

Intuitively, our network first makes a coarse estimate of the classification of parts to which  the pixel belongs and then predicts the exact position through small-scale correction. The coordinate regression at an image position $i$ can be given as
\begin{equation}
\label{equ:densepose}
	n^{*}=\mathop{\arg\max}_{n}P(n|p_i), [U,V]=R^{n^{*}}(p_i),
\end{equation}
where $n^{*}$ means the $id$ of vehicle part and $R^{n^{*}}$ means the regressor that predicts the U,V coordinates. In the first step (classification branch), the network assigns the pixel $p_i$ to the body part $n^{*}$ with the highest posterior probability. In the second step, we use the regressor $R^{n^{*}}$  to predict the U, V coordinates of pixel $p_i$ in part $n^{*}$. As described in Section~\ref{sec:3D_car_representation}, our template 3D vehicle contains $18$ parts; hence, $n$ takes $19$ values (one is background), and we train $18$ regression functions $R^{n}$ to obtain the 2D $[U, V]$ coordinates within each part $n$. In the training process, a cross-entropy loss $\mathit{L_{part}}$  and a smooth loss $\mathit{L_{UV}}$ are used in the first step (classification branch) and second step (regression branch), respectively.

\subsection{Vehicle Type Classification}\label{sec:type_prediction}
As described in Sec.~\ref{sec:dataset}, our deformable template can represent many vehicle classes, including coupe, hatchback, notchback, SUV, MPV, etc. Exact vehicle type is a useful cue for optimizing the pose and shape, which is important for 3D reconstruction (Sec.~\ref{sec:shape_pose}). Thus, we add a sub-branch to predict the vehicle type using a cross-entropy loss $\mathit{L_{type}}$.

We train the first stage of our deep network using the dense mapping data. The multi-task loss is defined as:
\begin{equation}
\label{equ:first_stage}
\begin{split}
%L = L^{p}_{class} + L^{p}_{reg} + L^{r}_{class} + L^{r}_{box}\\
%+ L^{r}_{mask} + L^{r}_{part} + L^{r}_{UV} + L^{r}_{type},
L = L_{mask} + L_{part} + L_{UV} + L_{type},
\end{split}
\end{equation}
%\modified{where $\mathit{(.)^{p}}$ and $\mathit{(.)^r}$ indicate RPN and RCNN, respectively. We minimize our loss function using the SGD optimization~\cite{SGD-op} (Done) WHAT IS SGD with a weight decay of 0.0001 and a momentum of 0.9. The learning rate is initially set to 0.002, and reduced by 0.1 for every 5 epochs.}

The loss function is optimized by the SGD optimization~\cite{SGD-op} in the training process with a weight decay of 0.0001 and a momentum of 0.9. The learning rate is initially set to 0.002 and reduced by 0.1 for every 5 epochs.

\subsection{Preliminary 3D Position Prediction}
\label{sec:keypoint_pose}
6-DoF pose estimation from a single view is a challenging task~\cite{3d-detection-ad:cvpr2019,3d-object-df:cvpr2019}.  The underlying difficulty comes from the fact that a pinhole camera cannot obtain the absolute 3D position due to the projective transformation. Therefore, in the second stage, we design a simple and effective network to further regress the 3D position (x,y,z) for each detected vehicle. We pre-define $400$ key-points on a 3D vehicle template and formulate a connectivity graph. We associate the pixels of the detected vehicle with this connectivity graph. In particular, the predicted pixels' UV coordinates with the part labels (obtained in part segmentation in the first stage) are used to project the 2D pixels back to the 3D space to identify the corresponding 3D key-points. Once we have the correspondence between the 2D image pixels and 3D key-points, we can associate the key-point with the 2D image pixel's attributes, such as pixel image coordinates and depth. 
Finally, the key-point graph is encoded by a $400\times400$ matrix, where only the visible key points have the associated values. The key-point graph matrix is fed into the network, which uses ResNet18 as the backbone and uses a fully connected convolution to learn the 3D positions of the detected vehicles using the $L_1$ loss.

% \modified{6-DoF pose estimation from single view is a challenging task in CV and CG. Difficulty comes from that pinhole camera cannot obtain absolute 3D positions due to its perspective imaging. Therefore, in the second stage of network, we further regress the 3D position (x,y,z) for each detected vehicle. Specifically, we pre-define 400 key-points and their connecting structure on the 3D vehicle template. The predicted pixels' UV coordinates with the part labels (in the first stage) are then projected to the 3D space to identify the corresponding 3D key-points. The structure of these key-points is encoded by a $400\times400$ matrix,  which maps the ground-truth 3D position (x,y,z) provided by our dense mapping dataset. While training, we adopt the ResNet18 architecture to learn the 3D positions using the $L_1$ loss.}

%%%%%%%%%%%%%%%%%%%%%%%%%%%%%%%%%%%%%%%%%%%%%%%%%%%%%%%%%%
\section{3D Reconstruction of Vehicles}
\label{sec:shape_pose} 
%%%%%%%%%%%%%%%%%%%%%%%%%%%%%%%%%%%%%%%%%%%%%%%%%%%%%%%%%%
Given the dense correspondences, vehicle types and initial vehicle positions obtained from the approach described in Sec.~\ref{sec:2d_parsing}, we compute the accurately fitted vehicle model $\mathcal{M}$ by jointly optimizing the vehicle's pose $\mathcal{T}\in\mathbb{SE}_3$ and deforming the shape of the vehicle model $\mathcal{M}_0$. We combine robust dense correspondences, vehicle types and initial vehicle positions from learning-based methods with classical optimization techniques and compute an accurate geometric model of the vehicle. In this section, we present the overall approach, which combines deep learning methods with model-based optimization methods.

%%%%%%%%%%%%%%%%%%%%%%%%%%%%%%%%%%%%%%%%%%%%%%%%%%%%%%%%%%
\subsection{6-DoF Pose and Shape Estimation}
%%%%%%%%%%%%%%%%%%%%%%%%%%%%%%%%%%%%%%%%%%%%%%%%%%%%%%%%%%

We use a triangle mesh to represent the vehile model $\mathcal{M}$ based on  the classical "active shape model" formulation~\cite{cootes1995active}. As mentioned in Section~\ref{sec:3D_car_representation}, we use PCA to obtain the mean position of the mesh's vertices as well as the principal modes of their relative displacement. The final template model consists of the base model $\mathcal{M}_0$ and $m$ principal component directions, $p_j$, and corresponding standard deviations $\delta_j$, where $1\leq j \leq m$. Any 3D model $\mathcal{M}$ can  be represented as a linear combination of $r$ principal components with geometric parameters $s$, as shown in Eq.~\ref{equ:pcamodel}.

%\begin{equation}
%    \mathcal{M}(s) = \mathcal{M}_0 + \sum_{k=1}^{r}s_k\delta_kp_k,
%	\label{equ:pcamodel}
%\end{equation}
%where $s_k$ is the weight of the $k_{th}$ principal component.
To evaluate the matching of the deformed model $\mathcal{M}$ with the input image, we use several terms to formulate an optimization problem. First, based on dense correspondences from Section~\ref{sec:dataset} and using initial vehicle positions ($\overline{T}$ obtained in Section~\ref{sec:keypoint_pose}) and vehicle types ($V_t$ obtained in Section~\ref{sec:part_uv}) as initialize parameters, we utilize every pair of 2D-3D mapping, e.g., the 3D vertex $v_i$ and the 2D $[U,V]$ coordinate $t_i$, to define the data term:
\begin{equation}
    E_c(\mathcal{T}, \mathcal{M}) = \sum_{i}\| t_i - \pi(v_i, R, T, V_t) \|^2 ,
	\label{equ:termCor}
\end{equation}
where $\pi(v)$ is the projection function, $R$ and $T$ are the pose parameters that need to be optimized. The vehicle types can indicate the types of vehicles, such as coupe, SUV, MPV. etc, which can be regarded as prior (vehicle width, height and length. etc.) in the optimization. As our network produces a part-based segmentation map $\mathcal{I}_{\mathcal{S}}$ with a silhouette with pixel-level accuracy, we can further constrain the estimation of $\mathcal{T}$ and $\mathcal{M}$ using the part-level silhouette term. We render the current model with current $\mathcal{T}$ into the image space to synthesize the 2D part-level silhouette image $\mathcal{I}_{\mathcal{M}}$. Next, we define the cost using the color space distance between those two images:
\begin{equation}
    E_s(\mathcal{T}, \mathcal{M}) = \sum_{x,y} \|\mathcal{I}_{\mathcal{S}}(x,y)  - \mathcal{I}_{\mathcal{M}}(x,y) \|^2.
	\label{equ:termSil}
\end{equation}
% We further use depth to form the geometry term as
%
%\begin{equation}
%    E_g(\mathcal{T}, \mathcal{M}) = \sum_{i\in V}(\mathcal{F}(v_i))^2 + \|v_i - %\mathcal{M}(v_i) \|^2 ,
%	\label{equ:termGeo}
%\end{equation}
To maintain the topology during the deformation, we utilize a geometric gradient to regularize the resulting mesh $\mathcal{M}$, which forms the smoothness term as:
\begin{equation}
    E_r(\mathcal{T}, \mathcal{M}) = \sum_{i}\|\nabla\mathcal{M}_0(v_i) -  \nabla\mathcal{M}(v_i)\|^2.
	\label{equ:termReg}
\end{equation}
Combining those three terms, we derive our optimization formulation as:
\begin{equation}
    %\min E(\mathcal{T}, \mathcal{M}) = \lambda_c E_c + \lambda_s E_s + \lambda_g E_g + \lambda_r E_r.
    \min E(\mathcal{T}, \mathcal{M}) = \lambda_c E_c + \lambda_s E_s + \lambda_r E_r.
	\label{equ:opt}
\end{equation}

The term weights in Eq.~\ref{equ:opt} are set to $\lambda_c = 1.0$, $\lambda_s=1.0$, and $\lambda_r=0.5$. We minimize Eq.~\ref{equ:opt} using an optimization solver, which is introduced in 2D/3D registration \cite{bouaziz2013dynamic}. Specifically, this minimization can be iteratively computed by two alternate steps. In the first step, a 3D rigid alignment is performed to estimate the 6-DoF pose. In the second step, a 2D/3D alignment of the deformable model is computed to optimize the shape parameters.
% WHAT KIND OF OPTIMIZATION METHOD DO YOU USE?
%Specifically, we run 100 iterations or when energy converge to target threshhold.

\begin{figure}
  \centering
  \includegraphics[width=1.0\linewidth]{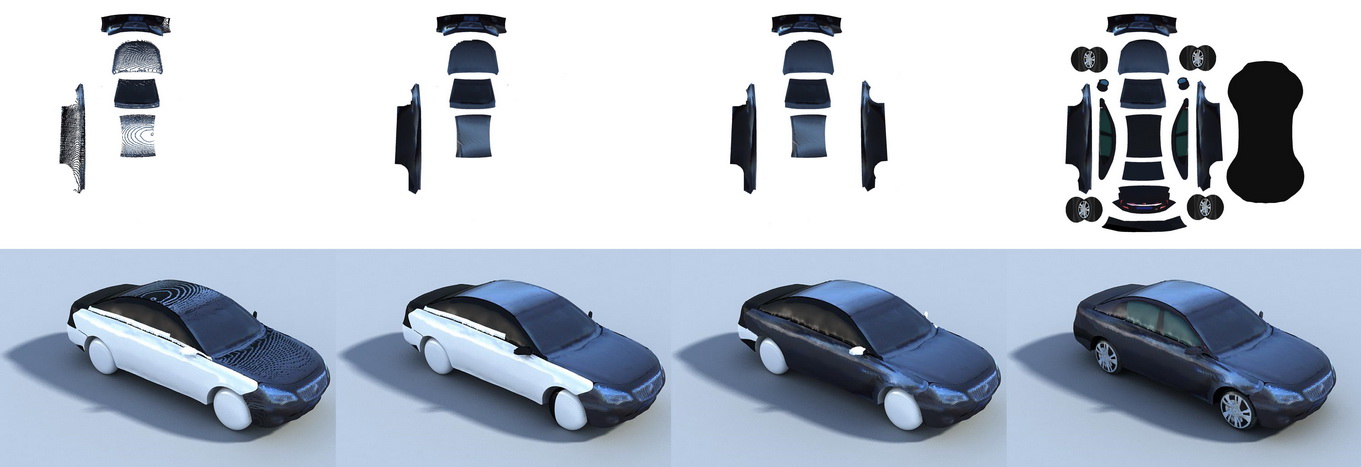}
  \caption{Intermediate results of appearance reconstruction: The top images are the reconstructed texture maps, and the bottom images are the corresponding rendered models in a different view. From left to right, we see the results of reconstructing from only visible pixels (a), with part-level completion (b), with symmetric completion (c), and with prior appearance completion (d).}
  \label{fig:TexturePipeline}
\end{figure}

\subsection{Appearance Extraction from Single View}
%In addition to pose and shape of the reconstructed vehicles, we also care about modeling appearance of vehicles. Technically, we extract realistic textures from the RGB images as the models are accurately aligned to the images. We solve the appearance extraction problem according to the method in~\cite{kholgade20143d}. Due to self occlusion and mutual occlusion, result models always have large hidden parts with missing texture. Thus, we further complete the appearance by extending method of~\cite{kholgade20143d}. Specifically, we first use predefined multiple planes of symmetry to complete the appearance in hidden parts of the object using the visible parts. After symmetric completion, for those still unsolved parts, we utilize appearance prior from the original texture of our PCA model $\mathcal{M}_0$ to inpaint the missing textures. Concretely, we solve such inpainting problem utilize gradient based image completion methods\cite{perez2003poisson}, where the gradients are defined by original texture of $\mathcal{M}_0$ and boundaries are grabbed from already reconstructed textures.
%, since models always have large hidden parts with missing texture due to self occlusion and mutual occlusion. 

In addition to the poses and shapes of vehicles, the appearance is also an important part of reconstruction. However, recovering the appearance, i.e. texturing the fully reconstructed vehicle model from a single image, is seen as an ill-posed problem \cite{wei2009state, huang2018appearance}. The key to solving such a problem is to tackle the appearance and shape priors of the vehicle model. Overall, the missing appearance of each vehicle part can be inferred by the original textures of the PCA model. Furthermore, the shape and appearance tend to be symmetric for vehicles. With those priors, we are able to texture the fully fitted model realistically with an extracted incomplete appearance from a single image. We use the algorithm described in~\cite{kholgade20143d,perez2003poisson} to solve the appearance extraction problem within a gradient-based image editing framework:
\begin{equation}
\begin{aligned}
    & min \sum_i^n \int_{\Omega_i} \| \nabla \mathcal{I_E}(t_i) - g_i\| ^2 
    \\ & s.t. \quad \ \mathcal{D}(\mathcal{I_E}(t_i), \mathcal{I}(\pi(v_i))), \quad \mathcal{S}(t_i, t_j),
    \label{tex}
\end{aligned}
\end{equation}
where $\Omega_i$ is the atlas area of the $i_{th}$ part in the texture map $\mathcal{I_E}$. $t_i\in\Omega_i$ is the 2D $[U,V]$ coordinate in the current atlas, and $v_i$ is the corresponding vertex of $t_i$ in the PCA model $\mathcal{M}_0$. Here, $g_i$ is the gradient of the original texture of $\mathcal{M}_0$. $\mathcal{D}(,)$ is the data constraint derived from the dense correspondence that keeps the color in the texture space the same as that in the input image $\mathcal{I}$. $\mathcal{S}(h_i, h_j)$ is the symmetry constraint defined by multiple planes, which is used to complete the appearance in hidden parts using the visible parts. The effectiveness of our appearance reconstruction is shown in Fig~\ref{fig:TexturePipeline}.

%Specifically,  After symmetric completion, for those still unsolved parts, we utilize appearance prior from the original texture of our PCA model  to inpaint the missing textures. Concretely, we solve such inpainting problem utilize gradient based image completion methods\cite{perez2003poisson}, where the gradients are defined by original texture of $\mathcal{M}_0$ and boundaries are grabbed from already reconstructed textures.

%Technically, we extract realistic textures from the RGB images as the models are accurately aligned to the images. We solve the appearance extraction problem according to the method in~\cite{kholgade20143d}. Thus, we further complete the appearance by extending method of~\cite{kholgade20143d}. Specifically, we first use predefined multiple planes of symmetry to complete the appearance in hidden parts of the object using the visible parts. After symmetric completion, for those still unsolved parts, we utilize appearance prior from the original texture of our PCA model $\mathcal{M}_0$ to inpaint the missing textures. Concretely, we solve such inpainting problem utilize gradient based image completion methods\cite{perez2003poisson}, where the gradients are defined by original texture of $\mathcal{M}_0$ and boundaries are grabbed from already reconstructed textures.

%%%%%%%%%%%%%%%%%%%%%%%%%%%%%%%%%%%%%%%%%%%%%%%%%%%%%%%%%%%%%%%%%

\section{Results and Discussion}
\label{sec:Experimental_Results}

%%%%%%%%%%%%%%%%%%%%%%%%%%%%%%%%%%%%%%%

In this section, we highlight the performance of our algorithms  on various public datasets that focus on Autonomous Driving Scenarios such as \textit{ApolloCar3D}~\cite{song2019apollocar3d}, \textit{CityScapes}~\cite{cordts2016cityscapes}, and \textit{KITTI}~\cite{geiger2012we}. To further demonstrate the robustness and generality of our framework, street-view images captured from mobile phones are also used for testing.

\textit{Manual Work:} While most of components of our approach are automatic, a few parts are generated with human or manual effort: 1) part-level vehicle 3D template construction and 2) data annotation of 6-DoF poses on KITTI dataset for ground-truth.
When the network has been trained, our approach is fully automatic without any human interactions  from the input (single image) to output (i.e. 2D detection, instance segmentation, part segmentation, U-V coordinate, 6-DoF pose, shape estimation, texture generation). Source code, data, and the trained model have been anonymously made public.

\begin{table}%
\begin{center}
\caption{Instance-level segmentation results of different methods (a higher value is better). Note that our method outperforms the SOTA method PANet by 6 points w.r.t. mAP. To be clear, only the ``Car'' category has been considered for evaluation on KITTI dataset.}
\begin{tabular}{l|ccc|ccc}
  \toprule
%  Methods      & $mAP$ & $AP_{50}$ & $AP_{75}$\\ \midrule
\multirow{2}{*}{Methods} & \multicolumn{3}{c|}{\textbf{ApolloCar3D}} & \multicolumn{3}{c}{\textbf{KITTI}}  \\  
    & $mAP$ & $AP_{50}$ & $AP_{75}$ & $mAP$ & $AP_{50}$ & $AP_{75}$              \\ \hline
  Mask R-CNN    & 39.1 & 68.1 & 41.4 & 41.5 & 68.7 & 47.2\\
  ApolloCar3D   & 42.2 & 61.0 & 43.9 & - & - & -\\
  PANet			& 42.4 & 73.1 & 44.8 & - & - & -\\
  Ours & \textbf{46.8} & \textbf{79.5} & \textbf{50.1} & \textbf{46.3} & \textbf{78.4} & \textbf{50.9} \\
  \bottomrule
\end{tabular}
%\vspace{0.1in}

\label{tab:instance_seg}
\end{center}
%\bigskip\centering
\end{table}%

%   Ours & \textbf{51.5} & \textbf{83.6} & \textbf{54.5} & \textbf{46.3} & \textbf{78.4} & \textbf{50.9} \\

%%%%%%%%%%%%%%%%%%%%%%%%
\subsection{Quantitative Evaluation}
\label{subsec:Quantitative_Evaluation}
%%%%%%%%%%%%%%%%%%%%%%%%
 %Currently, ApolloCar3D is the only public dataset that provides both 3D CAD models and car instance segmentation masks (reprojected by vehicles and corresponding poses). Thus, all the quantitative evaluations are conducted on the ApolloCar3D and the other public databases are only used for qualitative evaluations. Specifically, we conduct quantitative experiments on instance-level segmentation, 3D poses, and shape reconstruction. 

To evaluate our approach sufficiently, we conducted quantitative experiments on: \textit{1) instance-level segmentation}, \textit{2) 6-DoF pose estimation}, \textit{3) shape reconstruction}, and \textit{4) 3D detection} on various public benchmarks.

\subsubsection{Instance-level Segmentation}
%%%%%%%%%%%%%%%%%%%%%%%%%%%%%%%%%%%%%%%%%%%%%%%%%%%%%%%%%%%
%\subsubsection{Instance-level Segmentation}

%\paragraph{\textbf{Instance-level segmentation}} 

%\textbf{\modified{on Cityscape and ApolloCar3D}}

%\textit{\textbf{8.2.1 Instance-level Segmentation}}

Instance segmentation is regarded as a challenging problem in computer vision. We compare our method with three recent approaches: Mask-RCNN~\cite{maskrcnn:iccv2017}, a key-points based method~\cite{song2019apollocar3d}, and PANet~\cite{liu2018path} on the ApolloCar3D and KITTI datasets. PANet is the state-of-the-art published work of instance segmenta-tion for AD, which has the best performance for the “car” class on the CityScapes dataset.

For \textit{Mask R-CNN}, we use the code from GitHub implemented by Facebook AI Research~\footnote{https://github.com/facebookresearch/Detectron}. We adopt ResNet101~\cite{resnet-hekaiming:2015} as the backbone and follow the same training policy on the ApolloCar3D training dataset. Both are fine-tuned on the ApolloCar3D training dataset and the KITTI training dataset. 

In the \textit{ApolloCar3D} dataset, the car-fitting baseline methods provided are based on the key-points extraction. Based on the estimated pose and 3D car model, an instance-level mask can be easily generated by projecting the 3D model into the image coordinates.
%\modified{We would like to thank the authors of ApolloCar3D for providing us with the instance-level segmentation results directly.}

For \textit{PANet}, we use the code from GitHub implemented by authors~\footnote{https://github.com/ShuLiu1993/PANet} and re-train it on the ApolloCar3D dataset and the KITTI dataset.

All the evaluation results are given in Tab.~\ref{tab:instance_seg}. The method proposed in ApolloCar3D uses the 2D keypoints as training data, which are not provided by KITTI. Hence, we only compare our method with MaskRCNN on the KITTI dataset. From the table, we find that our method outperforms the other two methods by a large margin for all the evaluation criteria. For the ApolloCar3D dataset, compared with Mask-RCNN, ApolloCar3D, and PANet, the mAP provides improvements of \textit{7.7}, \textit{4.6}, and \textit{4.4} percentage points, respectively. In addition, for the KITTI dataset, compared with Mask-RCNN, the mAP has offered improvements of \textit{4.8}. Some instance segmentation results are shown in Appendix.  
% Fig~\ref{fig:seg-zoomin}. 
Note that heavily occluded vehicles are included in the ApolloCar3D and KITTI datasets, which makes it difficult to detect the masks based on the baseline Mask-RCNN approach. Additionally, the heavy occlusions also make it difficult to obtain accurate and stable key-points in dense traffic. Hence, the results obtained by ApolloCar3D are less stable and less accurate. Compared with these methods, our method obtains tighter and more accurate instance masks through part-level instance segmentation. Fig.~\ref{fig:seg-zoomin} highlights in detailed the differences produced by using different approaches.

\begin{table}[t]
\centering
\caption{6-DoF pose evaluation with different approaches. %where $^{*}$ means our in-house implementation.
$ApolloCar3D^{1}$ and $ApolloCar3D^{2}$  are the two improved versions in ApolloCar3D: ``Direct'' and  ``Key-points'', respectively. In addition, ``c-l'' indicates results from a loose criterion, and ``c-s'' indicates results from a strict criterion. ``A3DP-Abs'' and ``A3DP-Rel'' mean an absolute distance criterion and a relative distance criterion in evaluating transformation parameters of vehicle poses. Details about the evaluation criteria can be found in the Appendix. A higher value denotes better results.}
%\setlength\tabcolsep{7pt}
%\fontsize{8.5}{9}\selectfont
\begin{tabular}{l|lll|lll} 
\toprule[1pt]
\multirow{2}{*}{Methods} & \multicolumn{3}{c|}{\textbf{A3DP-Abs}} & \multicolumn{3}{c}{\textbf{A3DP-Rel}}  \\
                         & mean & c-l & c-s              & mean & c-l & c-s              \\ 
\hline
3D-RCNN& $16.44$  & $29.70$ & $19.80$ &$10.79$ & $17.82$ & $11.88$                   \\
$ApolloCar3D^{1}$ & $17.52$  & $30.69$ & $20.79$ &$13.66$ & $19.80$ & $13.86$ \\
\hline
DeepMANTA &$20.10$ & $30.69$ & $23.76$ & $16.04$ & $23.76$& $19.80$  \\
$ApolloCar3D^{2}$ & ${21.57}$ & ${32.62}$ & ${26.73}$ & ${17.52}$ & ${26.73}$ & ${20.79}$             \\ \hline
DensePose &$23.86$ & $39.60$ & $29.70$ & $17.62$ & $28.71$& $21.78$  \\
%arXiv Paper & ${29.01}$ & ${50.50}$ & ${40.59}$ & ${23.66}$ & ${41.58}$ & ${32.67}$             \\ \hline
Our Method &  $\textbf{32.97}$ & $\textbf{52.48}$ & $\textbf{40.59}$ & $\textbf{27.33}$ & $\textbf{45.54}$ & $\textbf{34.65}$\\
\bottomrule[1pt]
\end{tabular}
%\vspace{0.05in}
\label{tab:pose_estimation}
\end{table}

%%%%%%%%%%%%%%%%%%%%%%%%%%%%%%%%%%%%%%%%%%%%%%%%%%%%%%%%%%
\subsubsection{6-DoF Pose Estimation}
%%%%%%%%%%%%%%%%%%%%%%%%%%%%%%%%%%%%%%%%%%%%%%%%%%%%%%%%%%
%\paragraph{\textbf{3D pose estimation}}
Based on the dense 2D/3D mapping, we can estimate the 6-DoF pose for each vehicle instance. Generally speaking, 6-DoF pose estimation methods can be categorized into three groups: 1) \textit{``Direct'' methods}, which regress the vehicle poses from the 2D image directly using deep CNN-based networks; 2) \textit{``Key-points''-based methods}, which extract the predefined key-points of each vehicle from the image first and then compute the 3D pose via a PnP solver based on the matching between the 2D image and the 3D CAD model; and 3) \textit{``Dense mapping''-based methods}, which regress the dense mapping between the 2D image observation and the 3D model surface first; then, the 6-DoF pose is solved with a joint-optimization framework by considering all the mapping pixels. 
We compare our method with several state-of-the-art methods, such as~\textit{3D-RCNN}~\cite{kundu20183d}, \textit{DeepMANTA}~\cite{chabot2017deep}, Densepose~\cite{DensePose:CVPR2018} and \textit{ApolloCar3D}. For ~\textit{3D-RCNN}~\cite{kundu20183d} and~\textit{ApolloCar3D}, due to a lack of publicly available source codes, we reported the performance numbers from ApolloCar3D directly. For Densepose~\cite{DensePose:CVPR2018}, we retrain their network using the same training data with ours to obtain the dense mapping and use a PnP solver to compute the 3D pose of each vehile. 3D-RCNN and DeepMANTA are two representative works for ``Direct''- and ``Key-points''-based methods, respectively. Two improved versions are presented in~\cite{song2019apollocar3d}. Note that our method and Densepose can be categorized as a ``dense'' method.

\begin{figure*}
 \includegraphics[width=0.98\linewidth]{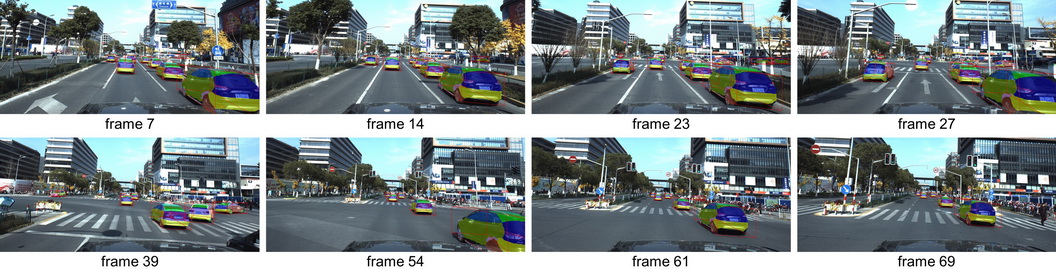}
%\caption{Segmentation examples in ApolloCar3D dataset for highlighting the advantage of our method to handle the occlusion cases.}
\caption{The figure shows the part-level segmentation and 3D detection results of the street view images. Note that these images are captured by a moving car, which is different from the training data (i.e. ApolloCar3D and KITTI). We obtain the results without network re-training or fine-tuning, which justifies the advantages of our approach. The results are included in the supplementary video.}
\label{fig:3d_detection_results}
\end{figure*}

The results of different approaches are shown in Tab.~\ref{tab:pose_estimation}, and we use ``A3DP-Abs'' and ``A3DP-Rel'' criteria as evaluation metrics, please see~\cite{song2019apollocar3d} for more details. From the table, it is easy to find that all the ``Key-points''-based and ``Dense''-based methods perform better than ``Direct''-based approaches. This is reasonable because prior knowledge (CAD model) of the object can help the pose estimation task. Besides, the ''Dense''-based approaches outperform ''Key-points''-based approaches, because dense 2D/3D mapping can provide more constrains, thus better results can be obtained. Comparing with all the methods, our methods outperforms other approaches with a large margin for both the ``A3DP-Abs'' and ``A3DP-Rel'' criteria. Comparing with ``Direct''-based methods, the mean relative and absolute precision have been improved by about \textit{15.0} percentage points with our method. For ``Key-points''-based methods, the mean relative and absolute precision have been increased by \textit{11.40} and \textit{9.81} percentage points, respectively.

For ''Dense mapping''-based approaches, after obtaining the dense 2D/3D mapping, a naive idea is to randomly sample some points and solve the pose by the PnP solver with our mean template vehicle model. We take this as our ``Dense'' baseline method (Densepose~\cite{DensePose:CVPR2018} shown in Table.~\ref{tab:pose_estimation}). To further refine the pose and the object shape, we take the pose estimated from the baseline as the initial value and refine it in conjunction with the model deformation. We represent this method as the ``Dense joint-optimization'' method in Tab.~\ref{tab:pose_estimation}. Comparing with Densepose~\cite{DensePose:CVPR2018}, our method can obtain vehicle types and initial vehicle positions in the process of network training, which are efficient and useful information to obtain better pose and shape in the PnP optimization process, thus better results can be obtained. Our experimental results show that based on the joint-optimization process, the mean precision has been improved by \textit{9.11} and \textit{9.71} percentage points for the ``A3DP-Abs'' and ``A3DP-Rel'' criteria, respectively.

\begin{table}
\begin{center}
\caption{Shape reconstruction errors by our approach. The lower the number, the higher the precision of the reconstructed model.}
\begin{tabular}{l|ccc}
  \toprule
%  Methods      & $mAP$ & $AP_{50}$ & $AP_{75}$\\ \midrule
     & Width & Height & Length      \\ \hline
  Mean Model of Ground Truth & 2.082m & 1.622m & 4.789m\\
  Mean Error of Reconstruction  & 0.050m & 0.079m & 0.129m \\
  Mean Error Rate  & \textbf{2.40$\%$} & \textbf{4.87$\%$} & \textbf{2.69$\%$} \\
  \bottomrule
\end{tabular}
%\vspace{0.05in}
\label{tab:recons_error}
\end{center}
%\bigskip\centering
\end{table}%

\begin{figure}
  \centering
  \includegraphics[width=0.95\linewidth]{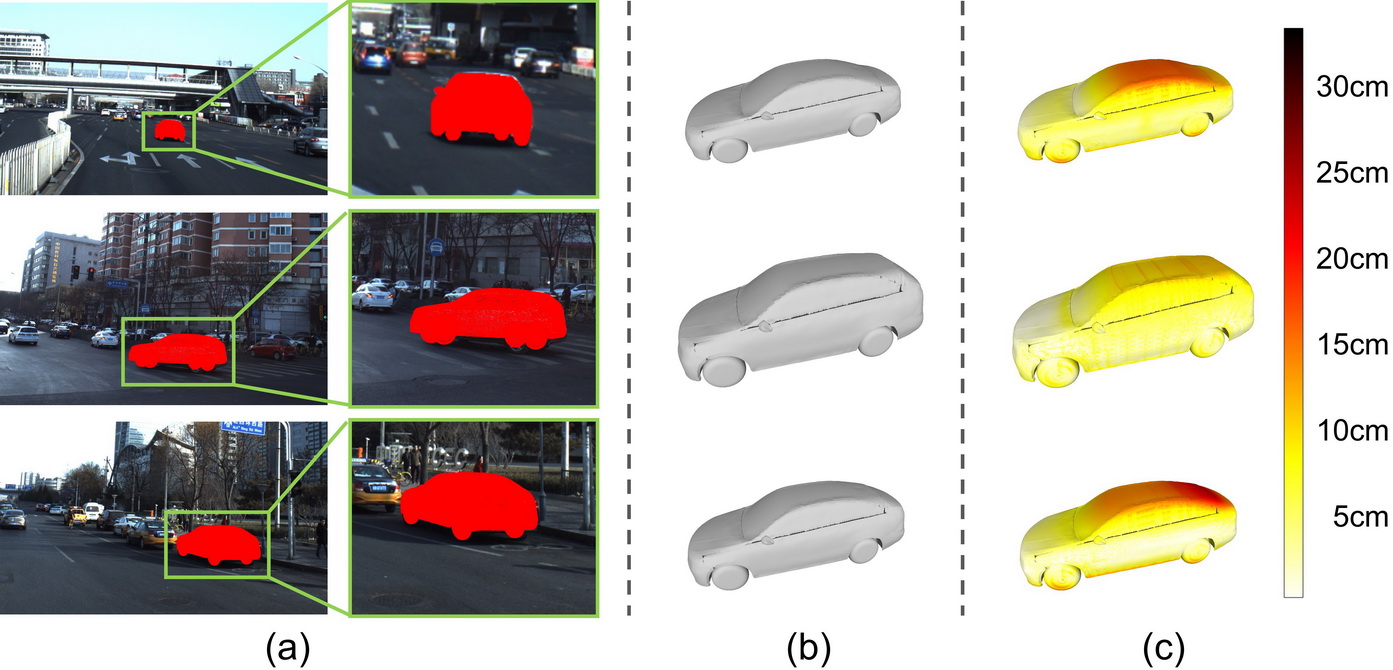}
  \caption{Shape estimation results from the real traffic images. (a) is the back-projection results from projecting the reconstructed shape model into the image coordinates. (b) is the reconstructed vehicle model and (c) is the error map compared with the ground truth model.}
  \label{fig:shape_reconstruction}
\end{figure}

%%%%%%%%%%%%%%%%%%%%%%%%%%%%%%%%%%%%%%%%%%%
\subsubsection{Shape Reconstruction}
%%%%%%%%%%%%%%%%%%%%%%%%%%%%%%%%%%%%%%%%%%%

We quantitatively evaluate the accuracy of our reconstructed 3D vehicle models on the validation dataset of ApolloCar3D, which consists of 200 street-view images with 2722 car instances. As shown in Tab.~\ref{tab:recons_error}, the mean error of reconstructed models is 0.050m (width), 0.079m (height), and 0.129m (length). Compared with the mean model ($2.082m \times 1.622m \times 4.789m$) of ground truth, the average reconstruction error is only \textit{2.40\%} (width), \textit{4.87\%} (height), and \textit{2.69\%} (length). Fig. \ref{fig:shape_reconstruction} illustrates some 3D reconstruction results for the real traffic scenario. From the reconstruction error distribution histograms, we can see that the reconstruction errors for most of the points are around 5$\sim$15cm. Fig. \ref{fig:shape_reconstruction} (a) shows the overlaid results by projecting the reconstructed 3D vehicle model into the image. We can see that the silhouette of the reconstructed model nearly coincides with the silhouette of the vehicle in the image.

%\modified{To evaluate accuracy of our reconstructed 3D vehicle models, we compare them with the ground truth. Fig.~\ref{fig:shape_reconstruction} illustrates some 3D reconstruction results for the real traffic scenario. From the reconstruction error distribution histograms, we can see that the reconstruction errors for most of the points are around 13 cm. Compared with a vehicle with 4 meter length, the reconstruction error is only 3\%. Fig \ref{fig:shape_reconstruction} (a) shows the overlaid results by projecting the reconstructed 3D vehicle model into the image. We can see that the silhouette of the reconstructed model nearly coincides with the silhouette of the vehicle in the image.} 

%%%%%%%%%%%%%%%%%%%%%%%%%%%%%%%%%
\subsubsection{3D Detection} 
%%%%%%%%%%%%%%%%%%%%%%%%%%%%%%%%%

We also evaluate our approach for 3D object detection on the KITTI dataset. This data has been divided into training and testing subsets, which consist of 7481 and 7518 frames, respectively. 
% Since the ground truth for the testing set is not available, we subdivide the training data into a training set and a validation set, as described in \cite{zhou2018voxelnet,zhou2019iou}. Finally, we obtain 3,712 data samples for training and 3,769 data samples for validation. 
On the KITTI benchmark, the objects have been categorized into ``easy,'' ``moderate,'' and ``hard'' based on their height in the image, occlusion ratio, etc. For each frame, both the camera image and the LiDAR point cloud have been provided, while only the color image from camera 2 has been used for our object detection. %The Lidar point cloud is only used for visualization.

On the KITTI benchmark, an average precision (AP) with Intersection over Union (IoU) that is larger than 0.7 is used as the
metric for evaluation. In practice, this is a very strict criterion for 3D object detection.  For fair comparison, we submitted the detection results to the official KITTI evaluation sever to obtain the quantitative number for the testing split. Several state-of-the-art monocular-based methods are used for the comparison, including FQNet~\cite{liu2019deep}, ROI-10D~\cite{manhardt2019roi}, GS3D~\cite{li2019gs3d}, MonoPSR~\cite{ku2019monocular}, and MultiFusion~\cite{xu2018multi}. As shown in Tab. \ref{tab:evaluation_with_other_methods_on_test_kitti}, our method outperforms all the other methods for all the ``easy',' ``moderate',' and ``hard'' categories.
% For the ``Easy'' category, the proposed method achieved the third place, which is is much betterthan FQNet, ROI-10D and GS3D.

\begin{table}[]
	\caption{ Comparison with other methods on the KITTI testing dataset for 3D ``Car'' detection. For clarity, we have highlighted the top  number in bold for each column. Higher numbers correspond to better results.}
		\begin{tabular}{r c ccc}
		\hline
		\multicolumn{1}{c }{\multirow{2}{*}{Methods}}& \multicolumn{1}{c}{\multirow{2}{*}{Modality}} & \multicolumn{3}{c}{\textbf{IoU = 0.7}} \\
		\multicolumn{1}{c}{}& \multicolumn{1}{c}{} & \multicolumn{1}{l}{Easy} & \multicolumn{1}{l}{Mod} & \multicolumn{1}{l}{Hard}\\ \hline
			FQNet~\cite{liu2019deep} &Mono &2.77 & 1.51 & 1.01 \\
			ROI-10D~\cite{manhardt2019roi} &Mono & 4.32 & 2.02 & 1.46 \\
			GS3D~\cite{li2019gs3d} &Mono & 4.47 & 2.90 & 2.47 \\
			MonoPSR~\cite{ku2019monocular} &Mono & {6.12} &4.00 &3.30 \\
%			TLNet~\cite{qin2019triangulation} &Stereo & 7.64 & 4.37 & 3.74 \\
%			MonoFENet~\cite{bao2019monofenet} &Mono & 8.35 & 5.14 & 4.10 \\
			MultiFusion~\cite{xu2018multi} &Mono &{7.08} & {5.18} & {4.68} \\
%			AM3D~\cite{ma2019accurate} &Mono & 16.50 &10.74  & 9.52 \\
%			D4LCN~\cite{ding2019learning} &Mono & 16.65 &11.72 & 9.51  \\
			Ours &Mono &{\textbf{8.45}} & \textbf{7.75} & \textbf{7.56}\\ \hline
		\end{tabular}
	\vspace{0.05in}
	\label{tab:evaluation_with_other_methods_on_test_kitti}
\end{table}

% \begin{table}%
% \begin{center}
% \begin{tabular}{l|ccc}
%   \toprule
% %  Methods      & $mAP$ & $AP_{50}$ & $AP_{75}$\\ \midrule
% \multirow{2}{*}{Methods} & \multicolumn{3}{c}{\textbf{KITTI}}  \\ 
%     & $mAP$ & $AP_{50}$ & $AP_{75}$   \\ \hline
%   Mask-RCNN~\cite{maskrcnn:iccv2017}    & - & - & -\\
%   \modified{PANet~\cite{liu2018path}}	& \modified{-} & \modified{-} & \modified{-} \\
%   Ours & \textbf{36.3} & \textbf{66.3} & \textbf{37.1} \\
%   \bottomrule
% \end{tabular}
% \vspace{0.05in}
% \caption{\modified{Instance segmentation results on the KITTI dataset. Note that all of three networks are trained on ApolloCar3D dataset (without KITTI data) and directly test on KITTI dataset. Results show that our method outperforms Mask-RCNN and PANet by $7$ and $5$ percentage points, respectively. It justifies the robustness and generability of our approach. }}
% \label{tab:without_training}
% \end{center}
% %\bigskip\centering
% \end{table}%

%%%%%%%%%%%%%%%%%%%%%%%%%%%%%%%%%%%%%%%%%%%%%%%%%%%%%%%%%
\subsection{Qualitative Evaluation}
\label{subsec:Qualitative_Evaluation}
%%%%%%%%%%%%%%%%%%%%%%%%%%%%%%%%%%%%%%%%%%%%%%%%%%%%%%%%%

%%%%%%%%%%%%%%%%%%%%%%%%%%%%%%%%%%%%%%%%%%%%%%%%%%%%%%%%%
\subsubsection{Generalizing Our Approach}
%%%%%%%%%%%%%%%%%%%%%%%%%%%%%%%%%%%%%%%%%%%%%%%%%%%%%%%%%

\begin{figure}
  \centering
  \includegraphics[width=0.98\linewidth]{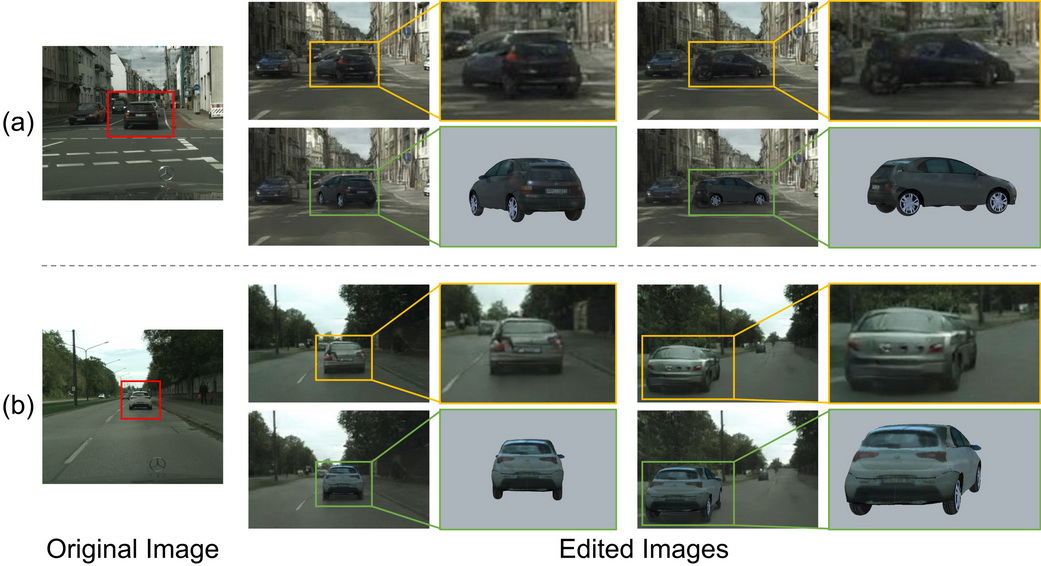}
  \caption{Image editing comparison of our method with Yao's method~\cite{yao20183d}: (a), (b) are images from the CityScapes dataset. Left column shows the original images with target vehicles highlighted in red boxes. Middle column and right column show images after vehicle translation and/or rotation. While there are obvious artifacts in Yao's results (yellow boxes), our method achieves better results visually (green boxes).}
  \label{fig:ImageEditing}
\end{figure}

Quantitative evaluation with other methods is important. However, for AD, robustness and generality are the most critical issues. Before we manually labeled the KITTI dataset, our deep network was only trained with the ApolloCar3D dataset. Next, this trained deep model is used for testing with other datasets directly without any ``fine-tuning'' or ``domain transfer'' strategies; these datasets include CityScapes, ApolloScape, and street-view images from an iPhone. The robustness and generality of our approach have been verified by these evaluations, which are shown in Fig. \ref{fig:3d_detection_results}, \ref{fig:ImageEditing}, % \ref{fig:comparison_with_mit_mask}, 
\ref{fig:ApolloSimulation},  \ref{fig:augmented_simulation},  \ref{fig:SeamlessMap}
%, ~\ref{fig:Simulation}  
%, \ref{fig:TemporalResults}
and Appendix. Note that the KITTI results in the supplementary files are without the trained dataset.

%Though our training images are only from ApolloCar3D, the trained network generalizes well on other datasets. Due to the lack of 3D car models in other datasets, we only show some qualitative results here. Fig.~\ref{fig:comparison_with_mit_mask} shows the instance segmentation comparison between our method and 3D-SDN \cite{yao20183d} (trained using vkitti dataset) on the Kitti dataset. While both methods are not trained on Kitti dataset, our method outperforms 3D-SDN~\cite{yao20183d} for the cases with occlusions and shadows. 

%%%%%%%%%%%%%%%%%%%%%%%%%%%%%%%%%%%%%%%%%%%%%%%%%%%%%%%%%
\subsubsection{Part-level Segmentation and 3D Detection on Video Sequence}
%%%%%%%%%%%%%%%%%%%%%%%%%%%%%%%%%%%%%%%%%%%%%%%%%%%%%%%%%

Fig.~\ref{fig:3d_detection_results} shows the part-level segmentation and 3D vehicle detection results from our approach in a sequence of street-view images. Different colors of cars in Fig.~\ref{fig:3d_detection_results} mean different parts segmented by our approach. It is clear that accurate and robust part-level segmentation results can be estimated by our network. Using the results obtained by our network, precise 6-DoF vehicle poses and 3D bounding boxes are calculated, which proves the effectiveness and accuracy of our approach in vehicle localization in autonomous driving. Note that these images are captured by a moving car, which is different from our training data (i.e. ApolloCar3D and KITTI). The results are generated without network ``re-training'' or ``fine-tuning,'' which justifies the advantages of our approach.

%%%%%%%%%%%%%%%%%%%%%%%%%%%%%%%%%%%%%%%%%%%%%%%%%%%%%%%%%
\subsubsection{Appearance Extraction for Image Editing}
%%%%%%%%%%%%%%%%%%%%%%%%%%%%%%%%%%%%%%%%%%%%%%%%%%%%%%%%%

In Fig.~\ref{fig:ImageEditing}, we also compare our method with the most recent ``inverse-graphics'' approach 3D-SDN~\cite{yao20183d}. Our approach benefits from having accurate poses and shapes, which means that we can extract realistic textures from real images. Thus, we gain the ability to edit images by rearranging the vehicles. This is achieved by removing the existing vehicles and inserting textured models back into the current image, which is also the main goal of 3D-SDN. From Fig.~\ref{fig:ImageEditing}, we can see that our method gets visually better results than 3D-SDN. Given the complete textured model, we can render consistent and visually compelling animation videos from different view-points using our approach (shown in the supplementary video), which is difficult for 3D-SDN to achieve.

%%%%%%%%%%%%%%%%%%%%%%%%%%%%%%%%%%%%%%%%%%%%%%%%%%%%%%%%%
\subsection{Performance Analysis}
%%%%%%%%%%%%%%%%%%%%%%%%%%%%%%%%%%%%%%%%%%%%%%%%%%%%%%%%%

% \subsubsection{The Impact of the Training Data}
% %%%%%%%%%%%%%%%%%%%%%%%%%%%%%%%%%%%%%%%%%%%%%%%%%%%%%%%%%
% \modified{x x x x x x x x x x x x x x x  x x x x x x x x x x x x x  x x x x x x x  x x x x x x x x  x x x x x  x x x x x  x x x }

\begin{figure}[!htbp]
  \centering
  \includegraphics[width=0.9\linewidth]{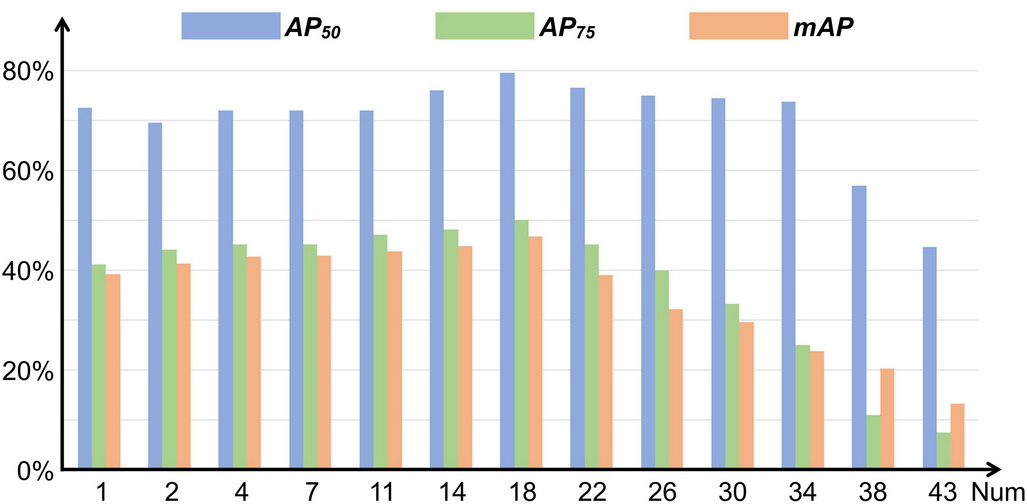}
  \caption{Instance-level segmentation results of different parts (vehicles separated from 1 part to 43 parts). X- and Y-axis represent the part number and $AP_{50}$, $AP_{75}$ and $mAP$ values,  respectively. All the values are the higher the better.}
  \label{fig:MultiPartsAP}
\end{figure}

%%%%%%%%%%%%%%%%%%%%%%%%%%%%%%%%%%%%%%%%%%%%%%%%%%%%%%%%%
\subsubsection{The Impact of the Part Number of Vehicle Template}
%%%%%%%%%%%%%%%%%%%%%%%%%%%%%%%%%%%%%%%%%%%%%%%%%%%%%%%%%

In addition, a series of experiments is also conducted to find an optimal number of parts to represent the vehicle model. The detailed comparison results are shown in Fig.~\ref{fig:MultiPartsAP}. First, the $mAP$ increases with the part number $N$ until it reaches 18. Then the mAP value drops dramatically with an increase in $N$. This can be explained because: appropriate parts representation can promote instance segmentation; however, too many parts may exceed the generalizability of part representation.

\begin{table}[t]
\centering
\caption{Ablation study of the 6-DoF pose evaluation with different approaches on ApolloCar3D benchmark. A, B and C indicate with Dense UV mapping, vehicle type, and initial vehicle 3D position, respectively.  %where $^{*}$ means our in-house implementation.
``c-l'' indicates results from a loose criterion, and ``c-s'' indicates results from a strict criterion. ``A3DP-Abs'' and ``A3DP-Rel'' mean an absolute distance criterion and a relative distance criterion in evaluating transformation parameters of vehicle poses. Details about the evaluation criteria can be found in the Appendix. A higher value denotes better results.}
%\setlength\tabcolsep{7pt}
%\fontsize{8.5}{9}\selectfont
\begin{tabular}{l|lll|lll} 
\toprule[1pt]
\multirow{2}{*}{Methods} & \multicolumn{3}{c|}{\textbf{A3DP-Abs}} & \multicolumn{3}{c}{\textbf{A3DP-Rel}}  \\
                         & mean & c-l & c-s              & mean & c-l & c-s              \\ 
\hline
A &$23.86$ & $39.60$ & $29.70$ & $17.62$ & $28.71$& $21.78$    \\
A + B & $28.71$  & $44.55$ & $34.65$ &$24.55$ & $38.61$ & $30.69$ \\
A + B + C &  $\textbf{32.97}$ & $\textbf{52.48}$ & $\textbf{40.59}$ & $\textbf{27.33}$ & $\textbf{45.54}$ & $\textbf{34.65}$\\
\bottomrule[1pt]
\end{tabular}
%\vspace{0.05in}

\label{tab:pose_estimation_ablation}
\end{table}

%%%%%%%%%%%%%%%%%%%%%%%%%%%%%%%%%%%%%%%%%%%%%%%%%%%%%%%%%
\subsubsection{Analysis 6-DoF Pose Estimation}
%%%%%%%%%%%%%%%%%%%%%%%%%%%%%%%%%%%%%%%%%%%%%%%%%%%%%%%%%

We provide more analysis of the effectiveness of the vehicle types obtained in Section~\ref{sec:part_uv} and vehicle positions obtained in Section~\ref{sec:keypoint_pose} in 6-DOF pose estimation. For results of Densepose~\cite{DensePose:CVPR2018}, after obtaining the dense 2D/3D mapping, we randomly sample some points and solve the pose by the PnP solver with our mean template vehicle model. The results are shown as the method A in Table.~\ref{tab:pose_estimation_ablation} . Vehicle types can provide useful prior information, such as the width, height and length of vehicles, and A $+$ B demonstrates the results of Densepose plus vehicle types which proves the effectiveness of vehicle types in pose estimation. It is easy to find that Densepose plus vehicle types can obtain more accurate 3D vehicle pose. Besides, initial vehicle positions can also help to improve the pose accuracy because it can provide rough positions of vehicles. A + B + C in Table.~\ref{tab:pose_estimation_ablation} demonstrates the pose result of Denpose plus vehicle types and vehicle initial positions, which outperforms Densepose and Densepose plus vehicle types, thus proves the effectiveness of using the learned initial positions to improve the overall accuracy of 6-DoF pose estimation.

\begin{figure}
  \centering
  \includegraphics[width=1.0\linewidth]{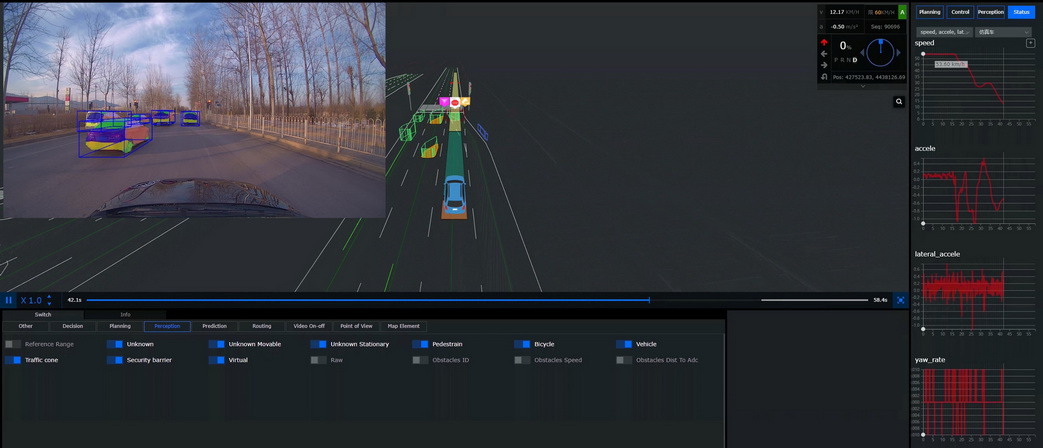}
  \caption{Results of integration of our approach with a vision-based autonomous driving system. The detected vehicle obstacles are then sent to the subsequent planner for trajectory generation.}
  \label{fig:ApolloSimulation}
\end{figure}

%%%%%%%%%%%%%%%%%%%%%%%%%%%%%%%%%%%%%%%%%%%%%%%%%%%%%%%%%%
%\subsection{Computation Time}
%%%%%%%%%%%%%%%%%%%%%%%%%%%%%%%%%%%%%%%%%%%%%%%%%%%%%%%%%%
%
%Our approach is optimized with CPU and GPU acceleration, and can be performed fully automatically without any human interactions. For a single image (3384 x 2710) containing more than 10 vehicles, the average processing time is about 0.32s, where 0.2s is for detection, segmentation and dense mapping, all of which are implemented using GPU, and 0.12s is for pose and shape estimation, which is multi-threaded. Note that 7s are needed for appearance reconstruction. In summary, for most autonomous driving applications without dependency on appearance, our approach could achieve a competitive processing rate at 3 fps. 

%%%%%%%%%%%%%%%%%%%%%%%%%%%%%%%%%%%%%%%%%%%%%%%%%%%%%%%%%
\subsection{Applications}
%%%%%%%%%%%%%%%%%%%%%%%%%%%%%%%%%%%%%%%%%%%%%%%%%%%%%%%%%
% DO NOT TALK ABOUT THE POTENTIAL OF ALL APPLICATIONS. RATHER ONLY THE ONES WHERE YOU SHOW ACTUAL RESULTS? THE REST OF THEM ARE FUTURE WORK.

Our approach detects, segments, and reconstructs textured 3D vehicle models from images. We highlight several applications of our approach in autonomous driving (AD) such as an AD perception pipeline, a data-driven AD simulator, AD training generation, and function analysis. The results of all applications are included in the supplementary video.

% The outputs of our approach can be easily applied in the following applications. 

\begin{figure*}
  \centering
  \includegraphics[width=0.94\linewidth]{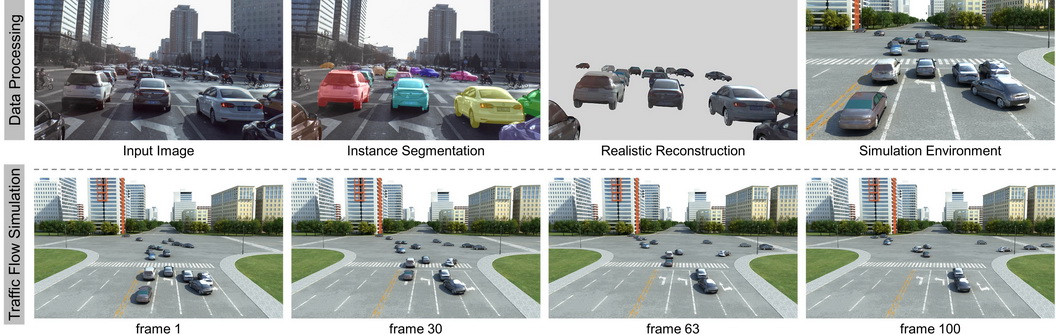}
  \caption{Traffic flow simulation results. The reconstructed 3D vehicles have been seamlessly transferred into a new environment. In addition, we can also design user-defined traffic flows by adding/deleting vehicles.}
  \label{fig:Simulation}
\end{figure*}
%%%%%%%%%%%%%%%%%%%%%%%%%%%%%%%%%%%%%%%%%%%%%%%%%%%%%%%%%
\subsubsection{AD Perception Pipeline}
%%%%%%%%%%%%%%%%%%%%%%%%%%%%%%%%%%%%%%%%%%%%%%%%%%%%%%%%%

Our approach can be used for vision-based or sensor-fusion-based perception pipelines. Fig. \ref{fig:ApolloSimulation} illustrates initial integration results for a real AD system. The ego-vehicle is equipped with multiple cameras, including front, left-side, and right-side cameras. Taking each raw camera image as input, our approach detects the surrounding vehicles and estimates their 3D poses and shapes. The detected vehicle obstacles are then sent to the subsequent planning and control module, which is responsible for generating vehicle control commands based on perceived objects. Compared to prior methods, our part-based approach generates more accurate 3D pose and shape representation, especially for heavy occlusion scenarios. More results are included in the supplementary video.

% , thus potentially improving the overall AD performance

%%%%%%%%%%%%%%%%%%%%%%%%%%%%%%%%%%%%%%%%%%%%%%%%%%%%%%%%%
\subsubsection{Data-driven Autonomous Driving Simulation}
%%%%%%%%%%%%%%%%%%%%%%%%%%%%%%%%%%%%%%%%%%%%%%%%%%%%%%%%%

\begin{figure}[h!]
  \centering
 \includegraphics[width=0.95\linewidth]{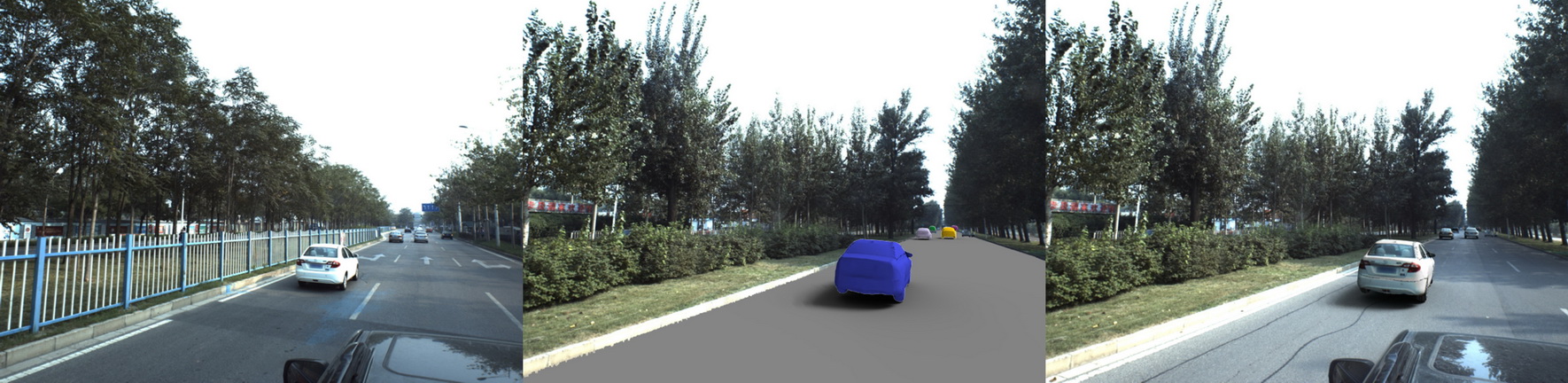}
\caption{Results of augmenting the autonomous driving simulator with the reconstructed realistic 3D vehicle models. The left image shows a captured camera image. The full reconstructed model (the white sedan) created with our approach can be used as a realistic asset in the simulator, which is shown in the right two images. }
\label{fig:augmented_simulation}
\end{figure}

%With the ability of generate 3D texture model 
The simulation system plays an essential role in autonomous driving. The augmented simulation~\cite{li2019aads}, which can build realistic scenes based on real elements from captured images, has recently attracted a great deal of attention. With our approach, information such as geometry models, appearance, and 6-DoF poses, can be completely reconstructed from a single image. This information can be seamlessly fed into the augmented simulation system to generate plausible simulation results. Fig.~\ref{fig:augmented_simulation} shows the augmented simulation result using our reconstructed 3D vehicle information.

The motion dynamics of real traffic are crucial to building a realistic simulator for testing decision and planning behaviors of a self-driving system. We further use our approach to capture the vehicle motions on video sequences. With simple tracking, we are able to capture motion dynamics in addition to 3D models of vehicles. As shown in Appendix,
% Fig.~\ref{fig:TemporalResults}
the reconstructed vehicle models in every frame are rendered and placed back into the point cloud of its original street level background. With accurate reconstruction in a single image, our approach could achieve temporally consistent results even without any temporal constraints, which can be seen from our supplemental video.

%%%%%%%%%%%%%%%%%%%%%%%%%%%%%%%%%%%%%%%%%%%%%%%%%%%%%%%%%
\subsubsection{Ground Truth Dataset Generation}

\begin{figure*}
  \centering
  \includegraphics[width=1.0\linewidth]{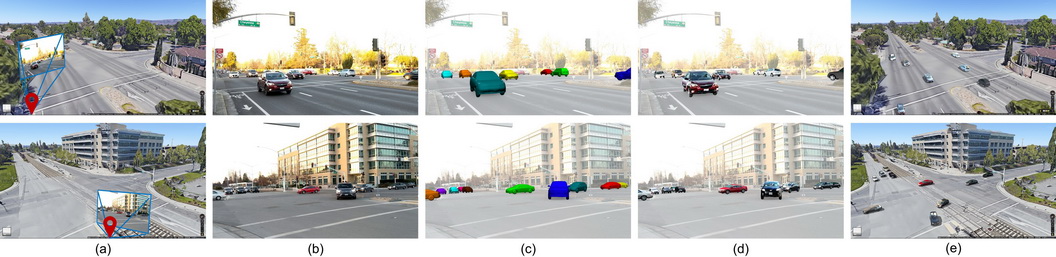}
  \caption{ Augmented Google 3D Maps with reconstructed vehicles: (a) The overhead views rendered from Google 3D Maps. Note that the red icons are positions where images in (b) are captured. (c) and (d) Images with instance segmentation and overlaid reconstructed vehicles, respectively. (e) Map images with augmented vehicles in the bird's eye view.}
  \label{fig:SeamlessMap}
\end{figure*}

Data labeling is time consuming and costly. The parsed 3D vehicle models by our approach can be re-rendered in either virtual or real scenarios to generate a large amount of ground truth data, which is urgently needed in autonomous driving for improving the deep neural network based vehicle detectors and trackers. In Appendix, we show the rendering of the reconstruction results for a new man-made CG environment. Fig.~\ref{fig:SeamlessMap} shows results of augmenting our reconstructed vehicle models into a reconstructed real environment (Google 3D map). We took pictures at random intersections in the bay area in California. Our algorithm is able to reconstruct 3D vehicle models of heavy traffic. We then overlay these models onto the same virtual locations within the Google 3D map.

%In the same way, we could also put our 3D textured vehicle models in the images of real scenes to generate some photo-realistic rendering results~\figref{fig:ImageEditing}.

%\paragraph{\textbf{Scene Re-composition}}
%Armed with complete vehicle information, we gain the ability to re-composite the captured scene by rearrange and render the reconstructed vehicles. \RED{Scene re-composition is hot application for "inverse-graphics" pipelines, such ours and ~\cite{yao20183d, kholgade20143d}. } 

%To achieve such goal, additional scene information(e.g. ground plane and normal, environment lighting) are needed. However, the interesting thing is that some of scene information can be naturally derived from our system. Concretely, as highly accurate vehicle 6-DoF poses are estimated, we could faithfully treat mean of the their upward axis as the ground normal. Furthermore, the lowest positions (ground-contacting point) of reconstructed vehicle models can be used for fitting the ground plane and detecting the drivable areas. With respect to lighting, we simplify the full environment lighting by the primary sun light, which could be queried from web\RED{[cite a paper or doc]} using capture location and time. With those information, we could rearrange vehicles and yield photo-realistic re-composited images.

%%%%%%%%%%%%%%%%%%%%%%%%%%%%%%%%%%%%%%%%%%%%%%%%%%%%%%%%%
\subsubsection{Function Analysis}
%%%%%%%%%%%%%%%%%%%%%%%%%%%%%%%%%%%%%%%%%%%%%%%%%%%%%%%%%
By parsing and reconstructing part-level vehicle information, our approach can be further used as a building block for high-level vision tasks such as function analysis and scene reasoning. 
For instance, based on our reconstructed part-level 3D models, we can directly edit the 2D images to generate a large amount of vehicles in `uncommon states' (\eg{}, door or trunk opening, headlights or taillight blinking). These data can be learned by a deep neural network to perform  holistically driving scene understanding \cite{liu2020parsing}. As shown in Fig. \ref{fig:Applications_AD}, the trunk of a car is detected as open with a human standing nearby, implying that the human may be taking the luggage out of the trunk. Thanks to the 6-DoF pose estimation and part-level 3D reconstruction of vehicles, we can further analyze these interactions in 3D space. As a response, the autonomous vehicle should slow down, turn the steering wheel, and change line.

\section{Conclusion, limitations and future work}\label{sec:Conclusion} 
In this paper we present a robust and effective approach to reconstruct complete 3D poses and shapes of vehicles from a single image. We introduce a novel part-level representation for vehicle segmentation and 3D reconstruction, which significantly improves the performance. Compared with state-of-the-art methods, our proposed middle part-level representation has several advantages: 1) based on the 2D/3D mapping, the 3D geometric information is implicitly embedded in the ground truth to guide the network learning; 2) the spatial relationship between different parts is taken into consideration for training, which is very useful in dealing with occlusion; and 3) our carefully designed parts definition and the PCA basis can be generalized and used for different types of vehicles. 

In addition to the part-based approach, we also generate the first 2D/3D dense mapping dataset and release this dataset. This new dataset, together with our new algorithm, leads to significantly better results in 3D vehicle detection and pose estimation.

\subsection*{Limitations and Future Work}
Although the benefit of our approach has been demonstrated in different scenarios, there are still some limitations: \begin{enumerate}[leftmargin=*]
    \item \textit{More 3D models:} Currently, the PCA basis is built on 78 CAD models. While this number of models can work well for most vehicle types, it fails for some special types such as pickup trucks, buses, or other road entities. To further enhance the represent ability of the PCA basis, more CAD models should be included in the model database. For urban traffic, we also need to be able to  reconstruct pedestrians and bicyclists.  

    \item \textit{More training data:} Our current model is only trained on the ApolloCar3D \cite{song2019apollocar3d} dataset and the KITTI \cite{geiger2012we} dataset. To improve the performance, we plan to use more datasets for training, such as the Cityscapes \cite{cordts2016cityscapes} or Virtual KITTI \cite{gaidon2016virtual} datasets. 
    \item \textit{Better texture handling:} We did not address the issue of lighting in the texture mapping process. Re-rendering the model with completely different lighting conditions may cause visual artifacts. We plan to add an environmental lighting estimation in the future. 
\end{enumerate}

%\section{Conclusion}
%The conclusion goes here.

% or


\appendices

%\section{}

\section{CAD Model and Template Alignment}
%%%%%%%%%%%%%%%%%%%%%%%%%%%%%%%%%%%%%%%%%%%
To build the PCA model, we have to first align the vehicle models with the average template model. Directly applying the non-rigid deformation for the whole model will cause some unexpected distortions on the tires. For example, the tires become elliptic rather than circular after deformation. To handle this, we divide model parts into two groups composed of four tires and other components and apply different deformation strategies for different groups.   

We first apply the ARAP (As-Rigid-As-Possible) deformation algorithm \cite{sorkine2007rigid} on all of the parts expect for the four tires to construct the deformation graph and align the target to the template. After the initial alignment, some coarse correspondences can be found. Then we non-rigidly deform the template body to the target model part with the correspondences found from the previous step. This bidirectional deformation schema can efficiently deform the template body to the target model. For the four tires, we apply a global alignment (\ie{} iterative closest point) algorithm to deform the template tires to the target model tires. After rotation, translation, and scaling operations, we finally assemble the deformed tires to the deformed body. In this way, we represent different car models with an average template model. From \figref{fig:PCA_Basis}, we see that our PCA basis has strong generalize ability and can produce both the SUV and the van, which are quite different from our template model.

\begin{figure*}
  \centering
  \includegraphics[width=0.98\linewidth]{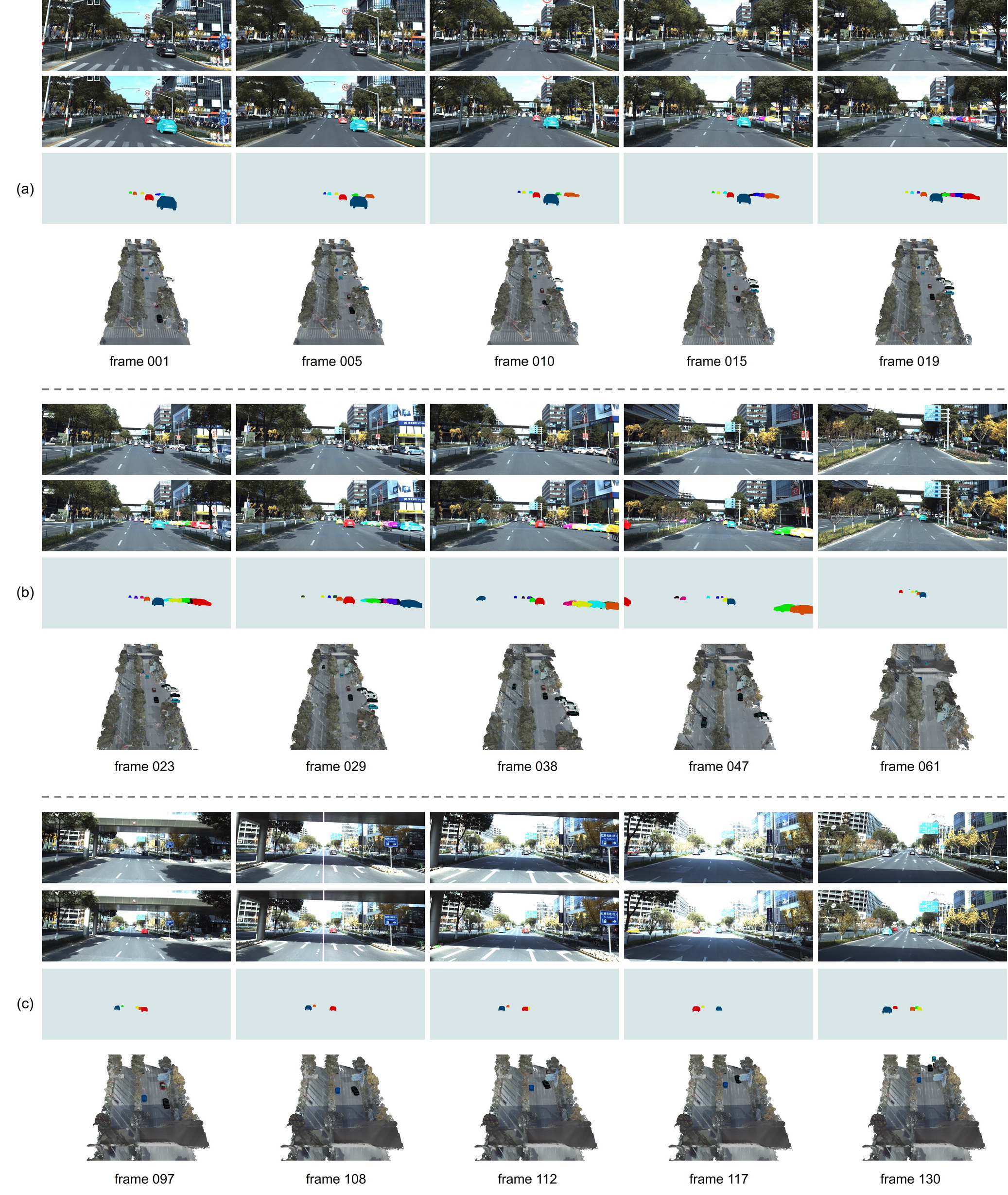}
  \caption{3D reconstruction results on an image sequence from the ApolloScape Dataset.}
 % \cite{wang2019apolloscape}.
  \label{fig:TemporalResults}
\end{figure*}

%%%%%%%%%%%%%%%%%%%%%%%%%%%%%%%%%%%%%%%%%%%
\section{Benchmarks for Evaluation}
%%%%%%%%%%%%%%%%%%%%%%%%%%%%%%%%%%%%%%%%%%%

\textit{CityScapes} has fine annotations for $2,975$ train, $500$ val, and $1,525$ test images. All images are 2048$x$1024 pixels. The instance segmentation task involves 8 object categories.

\textit{KITTI} is a large-scale dataset for autonomous diving scenarios collected from European streets. It provides various benchmarks for evaluation such as visual odometry, 2D/3D object detection and tracking, \etc For object detection tasks, it provides $7,481$ images for training and 7,518 images for testing.

\textit{ApolloCar3D} is a large-scale 3D instance car dataset built from real images captured in complex real-world driving scenes in multiple cities and targets 3D car understanding research in self-driving scenarios. For each car, the pose, 2D and 3D key-points such as corners of doors and headlights, instance level segmentation masks, and realistic 3D CAD models with an absolute scale are provided. Compared with CityScapes and KITTI, the traffic in ApolloCar3D is much more complex and includes more vehicles and more occlusion cases.

\textit{ApolloScape} is a large-scale AD-oriented dataset that provides video sequences in real street views (\figref{fig:TemporalResults}) 

\textit{Random street-view images.} To further evaluate our algorithm, we also collect some street-view images using mobile phones. Compared with images from the public dataset, the view-point and the image quality are quite different.   

%\begin{table}
%\begin{center}
%\begin{tabular}{l|ccc}
%  \toprule
%  
%  
%%  Methods      & $mAP$ & $AP_{50}$ & $AP_{75}$\\ \midrule
%     & mAP & AP50 & AP75      \\ \hline
%  without 33k instances & 36.7 & 60.2 & 41.9 \\
%  with 33K instances  & 46.3 & 78.4 & 50.9 \\
%  \bottomrule
%\end{tabular}
%\vspace{0.05in}
%\caption{According to the table, it is easy to find that instance segmentation results obtained with 33K vehicle instances for Kitti dataset outperforms results without such labels, which demonstrates that more training data will obtain better results.}
%\label{tab:label_33k}
%\end{center}
%%\bigskip\centering
%\end{table}%

%%%%%%%%%%%%%%%%%%%%%%%%%%%%%%%%%%%%%%%%%%%%%%%%%%%%%%%%%%%
%%%%%%%%%%%%%%%%%%%%%%%%%%%%%%%%%%%%%%%%%%%
\section{Evaluation Metric}
%%%%%%%%%%%%%%%%%%%%%%%%%%%%%%%%%%%%%%%%%%%

\textbf{\textit{Metric of Instance Segmentation.}} We use $mAP$, $AP_{50}$ and $AP_{75}$ as evaluative criteria for instance segmentation. Here, $mAP$ means the mean average precision under different IoU threshold between the prediction and ground-truth instance masks, where $AP_{50}$ and $AP_{75}$ mean the precision of IOU of $\geq 50\%$ and $\geq 75\%$, respectively. $mAP$, $AP_{50}$, and $AP_{75}$ are commonly used evaluative criteria. A detailed introduction of the evaluative criteria can be found in \cite{Everingham10}.

\textbf{\textit{Metric of 6-DoF Pose}.} 
We follow the evaluation criteria proposed in ApolloCar3D for 6-DoF pose evaluation, where \textit{``A3DP-Abs''} means the absolute distance criterion and \textit{``A3DP-Rel''} means the relative distance criterion. In addition, \textit{``c-l''} indicates results from a loose criterion and \textit{``c-s''} indicates results from a strict criterion. Specifically, in ApolloCar3D, the thresholds used for all levels of difficulty are $\{\delta_s\} = [0.5:0.05:0.95], \{\delta_t\} = [2.8:0.3:0.1], \{\delta_r\} = [\pi/6:\pi/60:\pi/60]$, where $[a:i:b]$ indicates a set of discrete thresholds sampled in a line space from $a$ to $b$ with an interval of $i$. In addition, the threshold for a loose criterion is $\ve{c}-l = [0.5, 2.8, \pi/6]$, while the  threshold for a strict criterion is $\ve{c}-l = [0.75, 1.4, \pi/12]$.

\ifCLASSOPTIONcaptionsoff
  \newpage
\fi



\bibliographystyle{IEEEtran}
\bibliography{bibliography}
\end{document}